%% file: BayesTucker.tex
\documentclass[10pt,journal,compsoc]{IEEEtran}
\pdfoutput=1

\usepackage{url}
\usepackage{hyperref}
\usepackage{color}
\usepackage{textcomp}
\usepackage{multirow}
\usepackage[cmex10]{amsmath}
\usepackage{amssymb}
\usepackage{stfloats}
\usepackage{amsthm}
\usepackage{bbm}
\usepackage[nocompress]{cite}
\usepackage[pdftex]{graphicx}
\graphicspath{{./images/}{./biography/} }
\DeclareGraphicsExtensions{.pdf,.jpeg,.png}
\usepackage[tight,normalsize,sf,SF]{subfigure}

\input{DefinedComand}

\begin{document}

\title{Bayesian Sparse Tucker Models for \\Dimension Reduction and Tensor Completion }

\author{Qibin Zhao,~\IEEEmembership{Member,~IEEE,}
        Liqing Zhang,~\IEEEmembership{Member,~IEEE,} and
        Andrzej Cichocki~\IEEEmembership{Fellow,~IEEE} % <-this % stops a space
\IEEEcompsocitemizethanks{\IEEEcompsocthanksitem Q. Zhao is with Laboratory for Advanced Brain Signal Processing, RIKEN Brain Science Institute, Japan and Department of Computer Science and Engineering, Shanghai Jiao Tong University, China.
\IEEEcompsocthanksitem L. Zhang is with MOE-Microsoft Laboratory for Intelligent Computing and Intelligent Systems and
Department of Computer Science and Engineering, Shanghai Jiao Tong University, China.
\IEEEcompsocthanksitem A. Cichocki is with Laboratory for Advanced Brain Signal Processing,
RIKEN Brain Science Institute, Japan and Systems Research Institute in Polish Academy of Science,  Warsaw,  Poland.
}
\thanks{} }

\markboth{Zhao \MakeLowercase{\textit{et al.}}: Bayesian Sparse Tucker Models for Dimension Reduction and Tensor Completion}%
{}

\IEEEcompsoctitleabstractindextext{%

\begin{abstract}
Tucker decomposition is the cornerstone of modern machine learning on tensorial data analysis, which have attracted considerable attention for multiway feature extraction, compressive sensing, and tensor completion.  The most challenging problem is related to determination of model complexity (i.e., multilinear rank), especially when noise and missing data are present. In addition, existing methods cannot take into account uncertainty information of latent factors, resulting in low generalization performance. To address these issues, we present a class of probabilistic generative Tucker models for tensor decomposition and completion with structural sparsity over multilinear latent space. To exploit structural sparse modeling, we introduce two group sparsity inducing priors by hierarchial representation of Laplace and Student-t distributions, which facilitates fully posterior inference. For model learning, we derived variational Bayesian inferences over all model (hyper)parameters, and developed efficient and scalable algorithms based on multilinear operations. Our methods can automatically adapt model complexity and infer an optimal multilinear rank by the principle of maximum lower bound of model evidence. Experimental results and comparisons on synthetic, chemometrics and neuroimaging data demonstrate remarkable performance of our models for recovering ground-truth of multilinear rank and missing entries.
\end{abstract}
}
\maketitle
\IEEEdisplaynotcompsoctitleabstractindextext
\IEEEpeerreviewmaketitle

\IEEEraisesectionheading{\section{Introduction}\label{sec:introduction}}
\IEEEPARstart{T}{ensor} decomposition aims to seek a multilinear latent representation of multiway arrays, which is a fundamental technique for modern machine learning on tensorial data mining. In contrast to matrix factorization, tensor decomposition can effectively capture higher order structural properties by modeling multilinear interactions among a group of low-dimensional latent factors, leading to significant advantages for multiway dimension reduction~\cite{zhao2013kernelization}, multiway regression/classification~\cite{zhao2013higher,zhao2013tensor}, and compressive sensing~\cite{caiafa2013computing} or completion of structured data~\cite{zhao2014bayesian}. In particular, Tucker~\cite{tucker1966some} and Candecomp/Parafac (CP)~\cite{bro1997parafac} decompositions have attracted considerable interest in various fields, such as chemometrics, computer vision, social network analysis, and brain imaging~\cite{Cichocki2009,kolda2009tensor,lu2011survey}.

Tucker decomposition represents an $N$th-order tensor by multilinear operations between $N$ factor matrices and a core tensor. If we view factor matrices as basis of $N$ latent space, then core tensor represents a multilinear coefficient for the whole tensor. An alternative perspective is that core tensor is viewed as basis of a latent manifold space, then factor matrices represent low-dimensional coefficients corresponding to $N$ different profiles for each data point. Tucker decomposition can exactly represent an arbitrary random tensor. In contrast, CP decomposition enforces a strict structure assumption by a super-diagonal core tensor, which is suitable for highly structured data but with limited generalization ability for various types of data. Tensor decomposition becomes interest only when underlying information has an intrinsic low-rank property, leading to a compact representation in latent space. Hence, one fundamental problem is separating the information and noise space by model selection, i.e., determination of tensor rank, which is challenging when noise variance is comparable to that of information. Note that tensor rank differs from matrix rank in many mathematic properties~\cite{de2008tensor,elizabeth2013tensor,lim2009most}.  Numerous studies proposed tensor decomposition methods by either optimization techniques or probabilistic model learning~\cite{Cichocki2009,kolda2009tensor,sorensen2012canonical, tensorlab2013,tao2008bayesian,gao2012probabilistic}. However, tensor rank is always required to be specified manually. In addition, general model selection methods need to perform tensor decomposition for each predefined model, which is inapplicable for Tucker model due to that the number of possibilities increases exponentially with the order of tensors. Some preliminary studies on automatic model determination were exploited in~\cite{morup2009automatic,yokota2014multilinear} which however, may prone to overffing and fail to recover true tensor rank in the case of highly noisy data, due to point estimation of model parameters.

Tensor completion can be formulated as tensor decomposition with missing values, which seems fairly straightforward but has several essential differences. Tensor decomposition focusing on data representation is referred to unsupervised technique, whereas tensor completion focusing on predictive ability is referred to semi-supervised technique, which aims to predict tensor elements from their $N$-tuple indices. Another important issue is that the optimal tensor rank for tensor completion is related to missing ratio and may differ from the underlying true rank that is always optimum for tensor decomposition. Hence, the key problem of tensor completion is to choose an appropriate tensor rank (i.e., model complexity) that can obtain optimal generalization performance, whereas this problem is more challenging as compared to tensor decomposition for a fully observed tensor. More specifically, existing model selections including cross-validation are not applicable due to correlations between the optimal tensor rank and number of observed data points. Although numerous tensor decomposition methods for partially observed tensor were developed in~\cite{filipovic2013tucker,chu2009probabilistic,xu2015bayesian,xiong2010temporal,acar2011scalable,hayashi2010exponential,jain2014provable}, the most challenging task of specifying tensor rank was conveyed to user, and thus was done mostly based on an implicit assumption that missing data is known. Several studies have considered automatic rank determination for CP decomposition~\cite{zhao2014bayesian,rai2014scalable}. In contrast to CP rank, multilinear rank has more degree of freedom. An alternative framework for tensor completion, based on a low-rank assumption, was developed by minimizing the nuclear norm of approximate tensor, which corresponds to a convex relaxation of rank minimization problem~\cite{liu2013tensorcompletion}. Although this technique successfully avoids specifying tensor rank manually, several tuning parameters are required and sensitive to missing ratio. Hence, it essentially transformed the model selection problem to a parameter selection problem. The nuclear norm based tensor completion was shown to be attractive in recent years~\cite{tan2014tensor,wang2014low,liu2014generalized,liu2014factor,signoretto2013learning,kressner2013low}. Many variants by imposing additional constraints were also exploited in~\cite{chen2013simul,narita2012tensor} which have shown advantages for some specific type of data. However, the best performance was mostly obtained by carefully tuning parameters  based on implicit assumption that missing data is known. Another issue is that the definition of nuclear norm of tensor corresponds to a (weighted) summation of mode-$n$ rank $R_n$ denoting the dimension of latent factor matrices. However, the dimension of core tensor $\prod_n R_n$ represents the model complexity of whole tensor as described previously. As a result, another possible framework can be introduced by optimization of logarithm transformed nuclear norm.

In this paper, we present a class of generative models for Tucker representation of a complete or incomplete tensor corrupted by noise, which can automatically adapt model complexity and infer an optimal multilinear rank solely from observed data. To achieve automatic model determination, we investigate structural sparse modeling through formulating Laplace as well as Student-t distributions in a hierarchial representation to facilitate full posterior inference, which therefore can be further extended to enforce group sparsity over factor matrices and structural sparsity over core tensor. For model learning, we derive the full posterior inference under a variational Bayesian framework, including remarkable inference for the non-conjugate hierarchical Laplace prior. Finally, all model parameters can be inferred as well as predictive distribution over missing data without needing of tuning parameters. In addition, we introduce several Theorems based on multilinear operations to improve computational efficiency and scalability.

The rest of this paper is organized as follows: Section~\ref{sec:notation} presents notations and multilinear operations. In Section~\ref{sec:sparsityprior}, we present group sparse modeling by hierarchical sparsity inducing priors. Section~\ref{sec:BTD} presents Bayesian Tucker model for tensor decomposition, while Section \ref{sec:BTC} presents Bayesian Tucker model for tensor completion. The algorithm related issues are discussed in Section \ref{sec:algorithm}. Section~\ref{sec:experiment} shows experimental results followed by conclusion in Section~\ref{sec:conclusion}.

\section{Preliminaries and Notations}
\label{sec:notation}
Let $\tensor{X}, \mathbf{X}, \mathbf{x}$ denote a tensor, matrix, vector respectively. Given an $N$th-order tensor $\tensor{X}\in\mathbb{R}^{I_1\times I_2\times \cdots \times I_N}$, its $(i_1, \ldots, i_N)$th entry is denoted by $\mathcal{X}_{i_1\cdots i_N}$, where $i_n=1,\ldots, I_n, n=1,\ldots, N$. The standard Tucker decomposition is defined by
\begin{equation}
\label{Eq:TuckerDecomposition}
\tensor{X} = \tensor{G}\times_1 \mathbf{U}^{(1)}\times_2\mathbf{U}^{(2)}\times\cdots\times_N\mathbf{U}^{(N)}.
\end{equation}
$\big\{\mathbf{U}^{(n)}\in\mathbb{R}^{I_n\times R_n}\big\}_{n=1}^{N}$ are a set of mode-$n$ factor matrices, $\tensor{G}\in\mathbb{R}^{R_1\times R_2\times\cdots\times R_N}$ denotes the core tensor and $(R_1,\ldots, R_N)$ denote the dimensions of mode-$n$ latent space, respectively. The overall model complexity can be represented by $\prod_n R_n$ or $\sum_n R_n$, whose minimum associated values $\{R_n\}_{n=1}^N$ is termed as \emph{multilinear rank} of tensor $\tensor{X}$~\cite{kolda2009tensor}. For a specific $\mathbf{U}^{(n)}$, we denote its row vectors by $\big\{\mathbf{u}^{(n)}_{i_n}\big\vert i_n=1,\ldots, I_n\big\}$ and its column vectors by $\big\{\mathbf{u}^{(n)}_{\cdot r_n}\big\vert r_n=1,\ldots, R_n\big\}$.

\begin{definition}
Let $\big\{\mathbf{U}^{(n)}\in\mathbb{R}^{I_n\times R_n}\big\}_{n=1}^N$ denote a set of matrices, the sequential \emph{Kronecker} products in a reversed order is defined and denoted by
\begin{equation}
\nonumber
\begin{split}
\bigotimes_n \mathbf{U}^{(n)}  &= \mathbf{U}^{(N)}\otimes \mathbf{U}^{(N-1)}\otimes\cdots\otimes \mathbf{U}^{(1)}.\\
\bigotimes_{k\neq n} \mathbf{U}^{(k)}  &= \mathbf{U}^{(N)}\otimes\cdots\otimes \mathbf{U}^{(n+1)}\otimes\mathbf{U}^{(n-1)}\otimes\cdots\otimes \mathbf{U}^{(1)}.
\end{split}
\end{equation}
The symbol $\otimes$ denotes \emph{Kronecker} product.  $\bigotimes_n \mathbf{U}^{(n)}$ is a matrix of size $\left(\prod_n I_n \times \prod_n R_n\right)$.
\end{definition}

The Tucker decomposition (\ref{Eq:TuckerDecomposition}) can be also represented by using matrix, vector, or element-wise forms, given by
\begin{equation}
\label{Eq:TuckerDecomposition2}
\begin{split}
\mathbf{X}_{(n)} &= \mathbf{U}^{(n)}\mathbf{G}_{(n)}\Bigg(\bigotimes_{k\neq n}\mathbf{U}^{(k)T}\Bigg),\\
\text{vec}(\tensor{X}) &= \bigg(\bigotimes_n \mathbf{U}^{(n)} \bigg)\mbox{vec}(\tensor{G}),\\
\mathcal{X}_{i_1 \cdots i_N} &= \bigg(\bigotimes_n \mathbf{u}^{(n)T}_{i_n} \bigg)\mbox{vec}(\tensor{G}).
\end{split}
\end{equation}
It should be noted that the multilinear operation is significantly efficient for computation. For example, if we compute $\bigotimes_n \mathbf{U}^{(n)}$ firstly and then multiply it with $\mathrm{vec}(\tensor{G})$, both the computation and memory complexity is $\mathcal{O}\left(\prod_n I_nR_n\right)$. In contrast, if we apply a sequence of multilinear operations $(\cdot)\times_n \mathbf{U}^{(n)}$ without explicitly computing $\bigotimes_n \mathbf{U}^{(n)}$, the computational complexity is $\mathcal{O}\left(\min_n (R_n)\prod_n I_n\right)$ while the memory cost is $\mathcal{O}(\prod_n I_n)$. In this paper, we use  notation $\bigotimes_n (\cdot)$ frequently for clarity, however, the implementation can be performed by using multilinear operations.

\section{Hierarchical Group Sparsity Priors}
\label{sec:sparsityprior}
The sparsity inducing priors are considerably important and powerful for many machine learning models. The most popular ones are Laplace, Student-t, and Spike and slab priors. However, these priors are often not conjugate with the likelihood distribution, which leads to difficulties for fully Bayesian inference. In contrast, another popular sparsity inducting prior is automatic relevance determination (ARD), which has been widely applied in many powerful methods such as relevance vector machine and Bayesian principle component analysis~\cite{tipping2001sparse,babacan2012sparse,bishop1999variational}. The advantages of ARD prior lie in its conjugacy resulted from a hierarchical structure. Note that ARD prior is essentially a hierarchical Student-t distribution since its marginal distribution is
\begin{equation}
\label{eq:HierarchialStudent}
\int_0^{\infty} \mathcal{N}(x|0, \lambda^{-1})Ga(\lambda|a,b)\, d\lambda = \mathcal{T}(x|0, ab^{-1},2a).
\end{equation}
$Ga(x|a, b) = \frac{b^a}{\Gamma{(a)}} x^{a-1} e^{-bx}$ denotes a Gamma distribution and $\mathcal{T}(x)$ denotes a Student-t distribution. One can show that when a noninformative prior is specified by $a=b\rightarrow 0$, then $p(x)\propto 1/|x|$, which indicates the sparsity inducing property of a hierarchial Student-t prior.

A straightforward question is that can we represent Laplace prior by the hierarchical distributions which are conjugate? As shown in~\cite{figueiredo2003adaptive,babacan2010bayesian}, a hierarchical structure of Gaussian and Exponential distributions yields a Laplace marginal distribution, whereas they are not conjugate priors.
In this paper, we present another hierarchial Laplace distribution by employing an Inverse Gamma distribution $IG(x|a,b) = \frac{b^a}{\Gamma{(a)}} x^{-a-1} e^{-bx^{-1}}$. By a specific setting of $IG(1, \frac{\gamma}{2}) =  \frac{\gamma}{2} x^{-2} e^{-\frac{\gamma}{2}x^{-1}}$, we can show that the marginal distribution is
\begin{equation}
\label{eq:HierarchialLaplace}
\int_0^{\infty} \mathcal{N}(x|0, \lambda^{-1})IG(\lambda|1,\frac{\gamma}{2})\, d\lambda = Laplace\big(x| 0, \frac{1}{\sqrt{\gamma}}\big).
\end{equation}
Note that $\gamma$ govern the degree of sparsity, for example, if $\gamma=1$, then $p(x)\propto e^{-|x|}$. To avoid parameter selection manually, we can place a noninformative prior over $\gamma$ and employ Bayesian inference for posterior estimation.  Although inverse Gamma is a non-conjugate prior, we can infer its variational posterior represented by a more generalized distribution.

Let $\mathbf{x}=\{x_1,\ldots, x_R\}$ denote a set of random variables, the hierarchical sparsity inducing priors can be specified as $\forall r=1,\ldots, R$,
\begin{equation}
\text{Student-t:} \quad x_r \sim \mathcal{N}(0, \lambda_r^{-1}), \quad \lambda_r \sim Ga(a,b),
\end{equation}
\begin{equation}
\text{Laplace:} \quad x_r \sim \mathcal{N}(0, \lambda_r^{-1}), \quad \lambda_r \sim IG(1,\frac{\gamma}{2}).
\end{equation}
Hence, the marginal distributions of $\mathbf{x}$ can be found by (\ref{eq:HierarchialStudent}), (\ref{eq:HierarchialLaplace}) as  $p(\mathbf{x}) = \prod_{r=1}^R \mathcal{T}(x_r|0, ab^{-1},2a)$ and $p(\mathbf{x}) = \prod_{r=1}^R Laplace(x_r|0, \frac{1}{\sqrt{\gamma}})$.

Now we consider to extend the above hierarchical priors for group sparse modeling. Let $\mathcal{X}=\{\mathbf{x}_1, \ldots, \mathbf{x}_R\}$ denote R groups of random variables where $\mathbf{x}_r \in R^{I_r}$ denote $r$th group that contains $I_r$ random variables. Note that $I_1=I_2\cdots=I_R$ is not necessary. The group sparse modeling is to enforce sparsity on groups in contrast to the individual random variables, which effectively take into account the clustering properties of relevant variables. To employ the hierarchial sparsity priors for group sparse model, it can be specified as $\forall r=1,\ldots, R$,
\begin{equation}
\text{Student-t:} \quad \mathbf{x}_r \sim \mathcal{N}(\mathbf{0}, \lambda_r^{-1}\mathbf{I}_{I_r}), \quad \lambda_r \sim Ga(a,b),
\end{equation}
\begin{equation}
\text{Laplace:} \quad \mathbf{x}_r \sim \mathcal{N}(\mathbf{0}, \lambda_r^{-1}\mathbf{I}_{I_r}), \quad \lambda_r \sim IG(1,\frac{\gamma}{2}).
\end{equation}
Therefore, the marginal distributions of $\mathcal{X}$ can be derived as $p(\mathcal{X})=\prod_{r=1}^R \mathcal{T}(\mathbf{x}_r|\mathbf{0}, ab^{-1}, 2a)$ and $p(\mathcal{X})=\prod_{r=1}^R Laplace(\mathbf{x}_r|\mathbf{0}, \frac{1}{\sqrt{\gamma}})$, where $\mathcal{T}(\mathbf{x}),\; Laplace(\mathbf{x})$ denote a multivariate Student-t distribution and a multivariate Laplace distribution respectively~\cite{babacan2014bayesian}. One can show that when $a=b\rightarrow 0$, then $\forall r, \; p(\mathbf{x}_r)\propto (1/\|\mathbf{x}_r\|_2)^{I_r}$. When $\gamma\rightarrow 1$, then $\forall r, \; p(\mathbf{x}_r) \propto e^{-\|\mathbf{x}_r\|_2}$.

In the following sections, we employ both hierarchial Student-t and hierarchial Laplace priors to impose group sparsity over multi-mode latent factors, yielding the minimum latent dimensions, for tensor decomposition models.

\section{Bayesian Tucker Model for Tensor Decomposition}
\label{sec:BTD}
\subsection{Model specification}
We first consider the Bayesian Tucker model for an $N$th-order tensor $\tensor{Y}\in\mathbb{R}^{I_1\times\cdots\times I_N}$ that is fully observed.
We assume that $\tensor{Y}$ is a measurement of the latent tensor $\tensor{X}$ corrupted by i.i.d. Gaussian noises, i.e., $\tensor{Y}=\tensor{X}+ \boldsymbol\varepsilon$, where $\tensor{X}$ is generated exactly by the Tucker representation as shown in (\ref{Eq:TuckerDecomposition}). Therefore, the observation model can be specified by a vectorized form,
\begin{equation}
\label{eq:BTDlikelihood}
{\small
\text{vec}(\tensor{Y})\Big| \!\left\{\mathbf{U}^{(n)}\right\}, \tensor{G}, \tau \;\sim\; \mathcal{N}\left(\bigg(\!\bigotimes_n \mathbf{U}^{(n)}\!\bigg) \text{vec}(\tensor{G}), \,\tau^{-1}\mathbf{I}\right).
}
\end{equation}
Our objective is to infer the model parameters, as well as  model complexity, automatically and solely from given data. To this end, we propose the hierarchical prior distributions over all model (hyper)parameters. For noise precision $\tau$, a \emph{Gamma} prior can be simply placed with appropriate hyperparameters, yielding an noninformative prior distribution.

To modeling tensor data by an appropriate model complexity, it is important to design the flexible prior distributions over the group of factor matrices $\{\mathbf{U}^{(n)}\}, n=1,\ldots, N$, and the core tensor $\tensor{G}$, whose complexity can be adapted to a specific given tensor. Since the model complexity of Tucker decompositions depends on the dimensions of $\tensor{G}$, denoted by $(R_1, R_2,\ldots, R_N)$, while $R_n, n=1,\ldots, N$ also corresponds to the number of columns in model-$n$ factor matrix $\mathbf{U}^{(n)}$ and thus represents the dimensions of mode-$n$ latent space. The minimal number of $\{R_n\}_{n=1}^{N}$ i.e., \emph{multilinear rank}~\cite{kolda2009tensor} usually need to be given in advance. However, due to the presence of noise, the optimal selection of multlinear rank is quite challenging. Although some model selection criterions can be applied, the accuracy significantly depends on the decomposition algorithms, resulting in less stability and high computational cost. Therefore, we seek an elegant automatic model selection, which can not only infer the multilinear rank, but also effectively avoid overfitting. To achieve this, we employ the proposed group sparsity priors over factor matrices. More specifically, each $\mathbf{U}^{(n)}$ is govern by hyperparameters $\boldsymbol\lambda^{(n)}\in \mathbb{R}^{R_n}$, where $\lambda_{r_n}^{(n)}$ controls the precision related to $r_n$ group (i.e., $r_n$th column). Due to the group sparsity inducing property, the dimensions of latent space will be enforced to be minimal. On the other hand, the core tensor $\tensor{G}$ also needs to be as sparse as possible. However, if we straightforwardly place an independent sparsity prior, the interactions between $\tensor{G}$ and $\{\mathbf{U}^{(n)}\}$ cannot be modeled, which may lead to inaccurate estimation of multilinear rank. As $\mathcal{G}_{r_1,\ldots, r_N}$ can be considered as the scalar coefficient of the rank-one tensor $\mathbf{u}^{(1)}_{\cdot r_1}\otimes\cdots\otimes\mathbf{u}^{(N)}_{\cdot r_N}$ that involves $r_n$th group from each of $\mathbf{U}^{(n)}$, respectively.  Taken into account the fact that if $\exists n, \exists r_n, \mathbf{u}_{\cdot r_n}^{(n)}=0$, then $\mathcal{G}_{r_1,\ldots, r_N}$ should be also enforced to be zero, the precision parameter for $\mathcal{G}_{r_1,\ldots, r_N}$ can be specified as the product of precisions over $\{\mathbf{u}_{\cdot r_n}^{(n)}\}_{n=1}^{N}$, which is represented by
\begin{equation}
\mathcal{G}_{r_1\cdots r_N} \Big| \left\{\boldsymbol\lambda^{(n)}\right\}, \beta \;\sim\; \mathcal{N}\left(0, \, \Big(\beta\prod_{n}\lambda_{r_n}^{(n)}\Big)^{-1}\right),
\end{equation}
where $\beta$ is a scale parameter related to the magnitude of $\tensor{G}$, on which a hyperprior can be placed. The hyperprior for $\boldsymbol\lambda^{(n)}$ play a key role for different sparsity inducing priors. We propose two hierarchial priors including Student-t and Laplace for group sparsity. Let $\boldsymbol\Lambda^{(n)} = \text{diag}(\boldsymbol\lambda^{(n)})$, we can finally specify the hierarchial model priors as
\begin{equation}
\label{eq:BTDprior}
\begin{split}
\tau \; &\sim\; Ga\big(a_0^\tau,\,b_0^\tau\big),\\
\text{vec}(\tensor{G}) \,\Big| \left\{\boldsymbol\lambda^{(n)}\right\}, \beta \;&\sim\; \mathcal{N}\left(\mathbf{0}, \, \left(\beta \bigotimes_n \boldsymbol\Lambda^{(n)}\right)^{-1}\right), \\
\beta \; & \sim\; Ga\big(a_0^\beta,\, b_0^\beta\big),\\
\mathbf{u}^{(n)}_{i_n}\big|\boldsymbol\lambda^{(n)} \; &\sim \; \mathcal{N}\left(\mathbf{0}, \, \boldsymbol\Lambda^{(n)^{-1}}\right), \quad \forall n, \forall i_n,\\
\text{Student-t:}\quad\lambda^{(n)}_{r_n} \; &\sim\; Ga\big(a_0^\lambda,\,b_0^\lambda\big), \quad  \forall n, \forall r_n,\\
\text{Laplace:}\quad  \lambda^{(n)}_{r_n} \; &\sim\; IG\big(1,\,\frac{\gamma}{2}\big), \quad  \forall n, \forall r_n,\\
 \gamma \; &\sim\; Ga(a_0^\gamma,\, b_0^\gamma).
\end{split}
\end{equation}
The prior for the core tensor $\tensor{G}$ is written as a tensor-variate Gaussian distribution. The observation model in (\ref{eq:BTDlikelihood}) and the hierarchial priors in (\ref{eq:BTDprior}) are integrated, which is termed as Bayesian Tucker Decomposition (BTD) model. Note that there are two proposed hierarchial sparsity priors in BTD model, which are thus denoted respectively by BTD-T (Student-t priors) and BTD-L (Laplace priors).

For simplicity, all unknown (hyper)parameters in BTD model are collected and denoted by $\Theta =\{\tensor{G},\mathbf{U}^{(1)}, \ldots, \mathbf{U}^{(N)}, \boldsymbol\lambda^{(1)},\ldots,\boldsymbol\lambda^{(N)}, \tau, \beta, \gamma\}$. Thus the joint distribution of BTD model is written as
\begin{multline}
\label{eq:jointdistribution}
p(\tensor{Y}, \Theta) = p\big(\tensor{Y}\vert\{\mathbf{U}^{(n)}\},\tensor{G},\tau\big) \prod_n p\big(\mathbf{U}^{(n)}\big\vert \boldsymbol\lambda^{(n)}\big)\\
\times p\big(\tensor{G}\big\vert \{\boldsymbol\Lambda^{(n)}\},\beta\big)\prod_n p\big(\boldsymbol\lambda^{(n)}|\gamma\big)p(\gamma)p(\beta)p(\tau).
\end{multline}
In general, maximum a posterior (MAP) of $\Theta$ can be estimated by optimizations of logarithm joint distribution w.r.t. each parameters alternately. However, due to the property of point estimation, MAP is still prone to overfitting. In contrast, we aim to infer the posterior distributions of $\Theta$ under a fully Bayesian treatment, which is $p(\Theta|\tensor{Y}) = \frac{p(\Theta, \tensor{Y})}{\int p(\Theta, \tensor{Y})\, d\Theta}$.

\subsection{Model inference}
The model learning can be performed by the approximate Bayesian inferences when the posterior distributions are analytically intractable. Since the variational Bayesian inference~\cite{winn2005variational} is more efficient and scalable as compared to sampling based inference methods, we employ VB technique to learn both BTD-T and BTD-L models. It should be noted that the hierarchial Laplace priors are not conjugate, resulting in a challenging problem to be addressed. In this section, we present the main solutions for model inference while the detailed derivations and proofs are provided in the Appendix.

VB inference aims to seek an optimal $q(\Theta)$ to approximate the true posterior distribution in the sense of $\min KL(q(\Theta)||p(\Theta|\tensor{Y}))$. Since $KL(q(\Theta)||p(\Theta|\tensor{Y})) = \ln p(\tensor{Y})- L(q)$, the optimum of $q(\Theta)$ can be achieved by maximization of lower bound $L(q)$ that can be computed explicitly.  To achieve this, we assume that the variational approximation posteriors can be factorized as
\begin{equation}
q(\Theta) = q(\tensor{G}) q(\beta)\prod_n q\big(\mathbf{U}^{(n)}\big)\prod_n q(\boldsymbol\lambda^{(n)}) q(\gamma)q(\tau).
\end{equation}
Then, the optimized form of $j$th parameters based on $\max_{q(\Theta_j)} L(q)$ is given by
\begin{equation}
q_j(\Theta_j) \propto \text{exp}\left\{ \mathbb{E}_{q(\Theta\backslash \Theta_j)} \left[\ln p(\tensor{Y}, \Theta)\right]\right\}.
\end{equation}
$\mathbb{E}_{q(\Theta\backslash\Theta_j)}[\cdot]$ denotes an expectation w.r.t. $q$ distribution over all variables in $\Theta$ except $\Theta_j$. In the following,  we use $\mathbb{E}[\cdot]$ to denote the expectation w.r.t. $q(\Theta)$ for simplicity.

As can be derived, the variational posterior distribution over the core tensor $\tensor{G}$ is
\begin{equation}
\label{eq:PostCore}
q(\tensor{G}) = \mathcal{N}\left(\text{vec}(\tensor{G})\big\vert \text{vec}(\widetilde{\tensor{G}}), \Sigma_G \right),
\end{equation}
where the posterior parameters can be updated by
{\small
\begin{gather}
\label{eq:InferMeanCore}
\text{vec}(\widetilde{\tensor{G}}) = \mathbb{E}[\tau] \, \Sigma_G \left(\bigotimes_n\mathbb{E}\left[\mathbf{U}^{(n)T}\right] \right)\text{vec}\left(\tensor{Y}\right),\\
\label{eq:InferVarCore}
\Sigma_G  =  \left\{ \mathbb{E}[\beta]\bigotimes_n \mathbb{E}\left[\boldsymbol\Lambda^{(n)}\right] + \mathbb{E}[\tau]\bigotimes_n \mathbb{E}\left[\mathbf{U}^{(n)T}\mathbf{U}^{(n)}\right]\right\}^{-1}.
\end{gather}
}Most of expectation terms in (\ref{eq:InferMeanCore}), (\ref{eq:InferVarCore}) are the functions linearly related to the corresponding $\Theta_j$, which can thus be easily evaluated from their posterior $q(\Theta_j)$. For example, $\mathbb{E}[\mathbf{U}^{(n)T}]$ can be evaluated according to $q(\mathbf{U}^{(n)})$ shown in (\ref{eq:postFactors}), as can be similarly computed for $\mathbb{E}[\tau]$, $\mathbb{E}[\beta]$, and $\mathbb{E}[\boldsymbol\Lambda^{(n)}]$. It should be noted that the expectation involving a quadratic term can be evaluated explicitly by using $\mathbf{E}[\mathbf{U}^{(n)T}\mathbf{U}^{(n)}]=\mathbb{E}[\mathbf{U}^{(n)T}]\mathbb{E}[\mathbf{U}^{(n)}]+ I_n \Psi^{(n)}$, which requires the posterior parameters given in (\ref{eq:postFactors}).

The computational complexity for posterior update of $\tensor{G}$ is $\mathcal{O}\left(\prod_n R_n^3 + \prod_n I_n R_n\right)$ that polynomially scales with model complexity denoted by $\prod_n R_n$ and linearly scales with data size denoted by $\prod_n I_n$. In general, it is dominated by $\mathcal{O}\left(\prod_n R_n^3\right)$, which is related to the matrix inverse. The memory cost is $\mathcal{O}\left(\prod_n R_n^2 + \prod_n I_n R_n\right)$, dominated by $\Sigma_G$ and $\otimes_n (\cdot)$. It should be noted that multilinear operations $\tensor{Y}\times_1 \mathbb{E}[\mathbf{U}^{(1)T}]\times\cdots\times_N \mathbb{E}[\mathbf{U}^{(N)T}]$ can be performed without explicitly computing $\otimes_n \mathbb{E}[\mathbf{U}^{(n)T}]$, thus reducing the memory cost to $\mathcal{O}\left(\prod_n R_n^2 + \prod_n I_n \right)$ and also reducing computation complexity to $\mathcal{O}(\prod_n R_n^3 + \prod_n I_n)$.  To further improve computation and memory efficiency, we introduce several important Theorems as follows.

\begin{theorem}
\label{theorem:1}
Let $\big\{\boldsymbol\Sigma^{(n)}\big\}$ be a set of diagonalizable matrices, and $c_1, c_2$ denote arbitrary scalars.
If $ \forall n=1, \ldots, N$,  the  spectral decompositions are represented by $\boldsymbol\Sigma^{(n)} = \mathbf{V}^{(n)}\mathbf{D}^{(n)}\mathbf{V}^{(n)T}$, then
{\small
\begin{multline}
\nonumber
\left(c_1\mathbf{I} + c_2\bigotimes_n \boldsymbol\Sigma^{(n)} \right)^{-1} =\\
\left(\bigotimes_n \mathbf{V}^{(n)} \right) \left(c_1\mathbf{I} + c_2\bigotimes_n \mathbf{D}^{(n)} \right)^{-1}\left(\bigotimes_n \mathbf{V}^{(n)T} \right).
\end{multline}
}
\begin{proof}
See Appendix for the detailed proof.
\end{proof}
\end{theorem}

\begin{theorem}
\label{theorem:2}
Let $\big\{\boldsymbol\Lambda^{(n)}\big\}$ be a set of diagonal matrices, $\big\{\boldsymbol\Sigma^{(n)}\big\}$ be a set of diagonalizable matrices, and $c_1, c_2$ be scalars. If $\forall n=1, \ldots, N$,  the spectral decompositions are represented by $\boldsymbol\Lambda^{(n)^{-\frac{1}{2}}}\boldsymbol\Sigma^{(n)}\boldsymbol\Lambda^{(n)^{-\frac{1}{2}}} = \mathbf{V}^{(n)}\mathbf{D}^{(n)}\mathbf{V}^{(n)T}$, then
{\small
\begin{multline}
\nonumber
\left(c_1\bigotimes_n \boldsymbol\Lambda^{(n)} + c_2\bigotimes_n \boldsymbol\Sigma^{(n)} \right)^{-1}=\\
\left(\bigotimes_n \boldsymbol\Lambda^{(n)^{-\frac{1}{2}}}\mathbf{V}^{(n)}\right)  \left(c_1\mathbf{I} + c_2\bigotimes_n \mathbf{D}^{(n)} \right)^{-1} \left(\bigotimes_n \mathbf{V}^{(n)T}\boldsymbol\Lambda^{(n)^{-\frac{1}{2}}}\right).
\end{multline}
}
\begin{proof}
See Appendix for the detailed proof.
\end{proof}
\end{theorem}

Based on Theorem \ref{theorem:1} and \ref{theorem:2}, $\Sigma_G$ can be factorized as the product of sequential Kronecker products, which leads to that matrix inverse operations can be performed by individual eigenvalue decompositions on $N$ small matrices, and inverse operations only on a diagonal matrix.  Therefore, the computational complexity for inference of $\tensor{G}$ can be significantly reduced to $\mathcal{O}\left(\sum_n R_n^3 + \prod_n I_n\right)$ while the memory cost can be significantly reduced to $\mathcal{O}(\sum_n R_n^2 + \prod_n I_n)$, if we save $\Sigma_G$ by a format of sequential Kronecker products.

As can be derived, the variational posterior distribution over the factor matrices $\big\{\mathbf{U}^{(n)}\big\}$ is represented by
\begin{equation}
\label{eq:postFactors}
q\big(\mathbf{U}^{(n)}\big) = \prod_{i_n=1}^{I_n} \mathcal{N}\left(\mathbf{u}_{i_n}^{(n)}\middle\vert \widetilde{\mathbf{\mathbf{u}}}^{(n)}_{i_n}, \Psi^{(n)} \right),\:   n = 1,\ldots,N,
\end{equation}
where the posterior parameters can be updated by
{\small
\begin{gather}
\label{eq:MeanFactors}
\widetilde{\mathbf{\mathbf{U}}}^{(n)} = \mathbb{E}[\tau]\,\mathbf{Y}_{(n)}\,\left(\bigotimes_{k\neq n}\mathbb{E}\left[\mathbf{U}^{(k)}\right]\right) \,\mathbb{E}\left[\mathbf{G}_{(n)}^T\right] \Psi^{(n)}, \\
\label{eq:VarFactors}
\Psi^{(n)} =  \left\{\mathbb{E}\big[\boldsymbol\Lambda^{(n)}\big] + \mathbb{E}[\tau] \mathbb{E}\left[\mathbf{G}_{(n)}\left(\bigotimes_{k\neq n} \mathbf{U}^{(k)T}\mathbf{U}^{(k)}\right)\mathbf{G}_{(n)}^T\right]\right\}^{-1}.
\end{gather}
}
In (\ref{eq:VarFactors}), the most complex expectation term related to multilinear operations can be computed by
{\small
\begin{multline}
\label{eq:ComputeFactor}
\text{vec}\left\{\mathbb{E}\left[\mathbf{G}_{(n)}\left(\bigotimes_{k\neq n}\mathbf{U}^{(k)T}\mathbf{U}^{(k)}\right)\mathbf{G}_{(n)}^T\right]\right\}\\
= \mathbb{E}\left[ \mathbf{G}_{(n)}\otimes \mathbf{G}_{(n)} \right]\text{vec}\left( \bigotimes_{k\neq n}\mathbb{E}\left[\mathbf{U}^{(k)T}\mathbf{U}^{(k)}\right]\right).
\end{multline}
}
Thus, each posterior expectation term can be evaluated easily according to the corresponding posterior distributions $q(\tensor{G})$ and $\{q(\mathbf{U}^{(k)})\}, k=1,\ldots, N, k\neq n$.

Taken into account the computation and memory efficiency, multilinear operations and sequential Kronecker products format must be employed to avoid explicitly computation of sequential Kronecker products. It should be noted that (\ref{eq:ComputeFactor}) cannot be factorized into operations on individual kronecker terms because of $\mathbb{E}\left[ \mathbf{G}_{(n)}\otimes \mathbf{G}_{(n)} \right]$. To reduce the memory cost, we may approximate it by $\mathbb{E}\left[ \mathbf{G}_{(n)}\right] \otimes \mathbb{E}\left[\mathbf{G}_{(n)} \right]$.  Therefore, the computational complexity for inference of $\mathbf{U}^{(n)}$ can be improved to $\mathcal{O}(R_n^3 + \prod_n I_n)$ while the memory cost is $\mathcal{O}\left(\prod_n R_n + \prod_n I_n\right)$.

The variational posterior distribution over $\beta$ can be derived to be a Gamma distribution due to its conjugate prior, which is denoted by $q(\beta)=Ga(a^\beta_M, b^\beta_M)$ where the posterior parameters can be updated by
\begin{equation}
\label{eq:PostBeta}
\begin{split}
a^\beta_M &= a^\beta_0 + \frac{1}{2}\prod_n R_n,\\
b^\beta_M &= b^\beta_0 + \frac{1}{2}\mathbb{E}\left[\text{vec}(\tensor{G}^2)^T\right]\bigotimes_n \mathbb{E}\left[\boldsymbol\lambda^{(n)}\right].
\end{split}
\end{equation}
In (\ref{eq:PostBeta}), $\mathbb{E}\left[\text{vec}(\tensor{G}^2)^T\right] = \text{vec}(\mathbb{E}[\tensor{G}]^2)^T + \text{diag}(\Sigma_G)^T$ should be applied for rigorous inference, whereas an alternative approximation is $\mathbb{E}\left[\text{vec}(\tensor{G}^2)^T\right] = \text{vec}(\mathbb{E}[\tensor{G}]^2)^T$ for efficiency.  $\{\mathbb{E}[\boldsymbol\lambda^{(n)}]\}$ can be easily evaluated according to $\{q(\boldsymbol\lambda^{(n)})\}$ described in the following paragraphs. The computational complexity in (\ref{eq:PostBeta}) is $\mathcal{O}(\prod_n R_n)$.

The inference of hyperparameters $\{\boldsymbol\lambda^{(n)}\}$ plays a key role for automatic model selection (i.e., determination of multilinear rank). In BTD models, as we proposed two hierarchical sparsity priors, resulting in two different posterior distributions for $\{\boldsymbol\lambda^{(n)}\}$.

\emph{BTD-T model using hierarchial Student-t priors.}
As can be derived, the variation posterior distribution over $\{\boldsymbol\lambda^{(n)}\}$ is i.i.d. Gamma distributions due to the conjugate priors, which is $\forall n=1,\ldots, N$,
\begin{equation}
\label{eq:PosterLambdaStudent}
q\big(\boldsymbol\lambda^{(n)}\big) = \prod_{r_n=1}^{R_n}Ga\big(\lambda^{(n)}_{r_n}\big\vert \tilde{a}_{r_n}^{(n)},\tilde{b}_{r_n}^{(n)}\big),
\end{equation}
where the posterior parameters can be updated by
\begin{equation}
\label{eq:InferLambda}
\begin{split}
\tilde{a}_{r_n}^{(n)} &= a_0^\lambda+ \frac{1}{2}\left(I_n + \prod_{k\neq n} R_k\right), \\
\tilde{b}_{r_n}^{(n)} &= b_0^\lambda+ \frac{1}{2} \mathbb{E}\left[\mathbf{u}^{(n)T}_{\cdot r_n}\mathbf{u}^{(n)}_{\cdot r_n}\right] \\
&\quad +\frac{1}{2}\mathbb{E}[\beta]\mathbb{E}\left[\text{vec}( \tensor{G}_{\cdots r_n\cdots}^2)^T\right]\bigotimes_{k\neq n}\mathbb{E}\left[\boldsymbol\lambda^{(k)}\right].
\end{split}
\end{equation}
According to $\{q(\mathbf{U}^{(n)})\}$ described in (\ref{eq:postFactors}), we obtain that
\begin{equation}
\mathbb{E}\big[\mathbf{u}^{(n)T}_{\cdot r_n}\mathbf{u}^{(n)}_{\cdot r_n} \big] = I_n \big(\Psi^{(n)}\big)_{r_n r_n} + \widetilde{\mathbf{u}}_{\cdot r_n}^{(n)T}\widetilde{\mathbf{u}}_{\cdot r_n}^{(n)}.
\end{equation}
This represents the posterior expectation of squared $L_2$-norm of $r$th component in mode-$n$ factors, which also takes into account the uncertainty information. $\mathbb{E}\left[\text{vec}( \tensor{G}_{\cdots r_n\cdots}^2)^T\right]$ represents the posterior expectation of squared $L_2$-norm of $r_n$ slice of core tensor $\tensor{G}$. Therefore, an intuitive interpretation of automatic model selection is that the smaller of $\big\|\mathbf{u}^{(n)}_{\cdot r}\big\|_F^2$ and $\|\tensor{G}_{\cdots r_n \cdots}\|_F^2$ leads to larger $\mathbb{E}[\lambda_{r_n}^{(n)}]$ and updated prior for $p\big(\mathbf{u}^{(n)}_{\cdot r_n}\big|\lambda_{r_n}^{(n)})$ as well as $p(\tensor{G}_{\cdots r_n\cdots}|\lambda_{r_n}^{(n)})$, which in turn enforces more strongly $\mathbf{u}^{(n)}_{\cdot r_n}$ and $\tensor{G}_{\cdots r_n\cdots}$ to be zero. After several iterations, the unnecessary columns in factor matrices and unnecessary slices in core tensor can be reduced to exact zero (i.e., smaller than machine precision).   The computational complexity for inference of $q(\boldsymbol\lambda^{(n)})$ is $\mathcal{O}(\prod_{n} R_n + I_n R_n)$. Given updated parameters for $q(\boldsymbol\lambda^{(n)})$, we can evaluate the posterior expectations by $\mathbb{E}[\boldsymbol\lambda^{(n)}] =\left[\tilde{a}_{1}^{(n)}/\tilde{b}_{1}^{(n)},\ldots,\tilde{a}_{R_n}^{(n)}/\tilde{b}_{R_n}^{(n)}\right]^T$ and $\mathbb{E}[\boldsymbol\Lambda^{(n)}] = \text{diag}(\mathbb{E}[\boldsymbol\lambda^{(n)}])$.

\emph{BTD-L model using hierarchial Laplace priors.} Since the hierarchical Laplace prior is not conjugate, which leads to much difficulties for model inference. To solve this problem, we employ a generalized inverse Gaussian distribution, denoted by $GIG(x|h, a, b)$, which includes Gamma, inverse Gamma, and  inverse Gaussian distribution as special cases by an appropriate setting of hyperparameters. For example, one can show that $Ga(a, b) = GIG(a, 2b, 0)$ and $IG(a, b) = GIG(-a, 0, 2b)$. For BTD-L mode, we keep using $\lambda^{(n)}_{r_n}\sim IG(1,\frac{\gamma}{2})$ as the hyper-prior, however the variational posterior $q(\lambda^{(n)}_{r_n})$ cannot be represented as an inverse Gamma distribution.

As can be derived, the variation posterior distribution over $\{\boldsymbol\lambda^{(n)}\}$ can be represented as i.i.d. GIG distributions, which is $\forall n=1,\ldots,N$,
\begin{equation}
\label{eq:PostLambdaLaplace}
q\big(\boldsymbol\lambda^{(n)}\big) = \prod_{r_n=1}^{R_n} GIG\big(\lambda^{(n)}_{r_n}\big\vert h, \tilde{a}_{r_n}^{(n)},\tilde{b}_{r_n}^{(n)}\big),
\end{equation}
where the posterior parameters can be updated by
\begin{equation}
\label{eq:InferLambdaLaplace}
\begin{split}
h &= \frac{1}{2}\left(I_n+\prod_{k\neq n}R_k\right) -1, \qquad \tilde{b}_{r_n}^{(n)} = \mathbb{E}[\gamma],\\
\tilde{a}_{r_n}^{(n)} &=  \mathbb{E}[\beta] \mathbb{E}\left[\text{vec}(\tensor{G}_{\cdots r_n\cdots}^2)^T\right]\bigotimes_{k\neq n}\mathbb{E}\left[\boldsymbol\lambda^{(k)}\right] \\
& \qquad + \mathbb{E}\left[\mathbf{u}^{(n)T}_{\cdot r_n}\mathbf{u}^{(n)}_{\cdot r_n}\right].
\end{split}
\end{equation}
The computational complexity for inference of $q(\boldsymbol\lambda^{(n)})$ is also $\mathcal{O}(\prod_n R_n + I_n R_n)$. Given the updated parameters, we can evaluate $\mathbb{E}_{GIG}\big[\lambda_{r_n}^{(n)}\big]$ straightforwardly, while an alternative approximation is the posterior mode w.r.t. GIG distribution that can avoid computational instabilities of modified Bessel function.

By comparing (\ref{eq:PostLambdaLaplace}) with (\ref{eq:PosterLambdaStudent}), we can investigate the essential difference between Student-t and Laplace priors. One can show that (\ref{eq:PosterLambdaStudent}) can be rewritten as $GIG\left(\tilde{a}_{r_n}^{(n)}, 2\tilde{b}_{r_n}^{(n)},0\right)$ with parameters given by (\ref{eq:InferLambda}). Hence, the key difference lies in the setting of $\gamma$. If $\gamma=0$, Student-t and Laplace priors are essentially equivalent. To avoid manually tuning parameters, we also place a hyper-prior over $\gamma$ and thus derive the variational posterior distribution as $q(\gamma) = Ga(a_M^\gamma, b_M^\gamma)$ whose posterior parameters can be updated by
\begin{equation}
\begin{split}
a_M^\gamma &= a_0^\gamma + \sum_{n=1}^N R_n, \\
b_M^\gamma &= b_0^\gamma + \frac{1}{2}\sum_{n=1}^N \sum_{r_n=1}^{R_n} \mathbb{E}\left[\lambda_{r_n}^{(n)^{-1}}\right].
\end{split}
\end{equation}
It should be noted that $\mathbb{E}\big[\lambda_{r_n}^{(n)^{-1}}\big]$ cannot be computed straightforwardly by $\mathbb{E}\big[\lambda_{r_n}^{(n)}\big]^{-1}$. Since $q(\lambda_{r_n}^{(n)})$ is a GIG distribution as shown in (\ref{eq:PostLambdaLaplace}), it is not difficult to derive that $q\big(\lambda_{r_n}^{(n)^{-1}}\big) = GIG(-h, \tilde{b}^{(n)}_{r_n}, \tilde{a}^{(n)}_{r_n})$, yielding the posterior expectation computed by
\begin{equation}
\mathbb{E}_{GIG}\left[\lambda_{r_n}^{(n)^{-1}}\right] = \frac{\sqrt{\tilde{a}^{(n)}_{r_n}}K_{1-h}\left(\sqrt{\tilde{a}^{(n)}_{r_n}\tilde{b}^{(n)}_{r_n}}\right)}{\sqrt{\tilde{b}^{(n)}_{r_n}} K_{-h}\left(\sqrt{\tilde{a}^{(n)}_{r_n}\tilde{b}^{(n)}_{r_n}}\right)},
\end{equation}
and the posterior mode computed by
{\small
\begin{multline}
\!\!\!\!\!\arg\!\!\max_{\lambda_{r_n}^{(n)^{-1}}}GIG\left(\lambda_{r_n}^{(n)^{-1}}\right) = \frac{(-h-1)+\sqrt{(-h-1)^2+\tilde{a}^{(n)}_{r_n}\tilde{b}^{(n)}_{r_n}}}{\tilde{b}^{(n)}_{r_n}},
\end{multline}
}where $K_{1-h}(\cdot)$ denotes a modified Bessel function of the second kind.

As can be derived, the variational posterior distribution over the noise hyperparameter is $q(\tau) = Ga(a_M^\tau, b_M^\tau)$ whose parameters can be updated by
{\small
\begin{equation}
\label{eq:PostTau}
\begin{split}
a_M^\tau &= a_0^\tau+\frac{1}{2}\prod_{n} I_n, \\
b_M^\tau &= b_0^\tau+\frac{1}{2} \mathbb{E}\left[\left\|\text{vec}(\tensor{Y})-\left(\bigotimes_n\mathbf{U}^{(n)}\right)\text{vec}(\tensor{G})\right\|_F^2\right],
\end{split}
\end{equation}
}where the posterior expectation of model residuals can be evaluated by
{\small
\begin{multline}
\label{eq:ModelError}
\mathbb{E}\left[\left\|\text{vec}(\tensor{Y})-\left(\bigotimes_n\mathbf{U}^{(n)}\right)\text{vec}(\tensor{G})\right\|_F^2\right] =\\ \|\tensor{Y}\|_F^2 -2\text{vec}(\tensor{Y})^T \left(\bigotimes_n\mathbb{E}[\mathbf{U}^{(n)}]\right)\mathbb{E}[\text{vec}(\tensor{G})] \\ + \text{Tr}\left(\mathbb{E}\left[\text{vec}(\tensor{G})\text{vec}(\tensor{G})^T \right]\bigotimes_n \mathbb{E}\left[\mathbf{U}^{(n)T}\mathbf{U}^{(n)}\right] \right).
\end{multline}
}In principle, $\mathbb{E}\left[\text{vec}(\tensor{G})\text{vec}(\tensor{G})^T \right] = \text{vec}(\widetilde{\tensor{G}})\text{vec}(\widetilde{\tensor{G}})^T + \Sigma_G$. However,  $\text{vec}(\widetilde{\tensor{G}})\text{vec}(\widetilde{\tensor{G}})^T $ can be alternatively used as an approximation, which then makes it possible to apply multilinear operations for computing (\ref{eq:ModelError}) quite efficiently. Hence, the computational complexity can then be reduced to $\mathcal{O}(\prod_n R_n + \prod_n I_n)$.

The inference framework presented in this section can essentially maximize the lower bound of model evidence which is defined by $\mathcal{L}(q) = \mathbb{E}_{q(\Theta)}[\ln p(\tensor{Y},\Theta)] + H(q(\Theta))$. The first term denotes the posterior expectation of joint distribution while the second term denotes the entropy of $q(\Theta)$. In principle, $L(q)$ should increase at each iteration, thus it can be used to test for convergence. We provide the detailed computation forms of $L(q)$ in the Appendix.

\section{Bayesian Tucker Model for Tensor Completion}
\label{sec:BTC}
\subsection{Model specification}
In this section, we consider Bayesian Tucker model for tensor completion. Let $\tensor{Y}$ denotes an incomplete tensor (i.e., with missing entries), and $\tensor{O}$ denotes a binary tensor indicating the observation positions, i.e., $\mathcal{O}_{i_1\cdots i_N} =1$ if $(i_1,\ldots, i_N)\in\Omega$ otherwise it is zero. $\Omega$ denotes a set of $N$-tuple indices of observed entries. $\tensor{Y}_\Omega$ denotes only observed entries. Similar to BTD model, we assume a generative model $\tensor{Y}_\Omega = \tensor{X}_\Omega + \varepsilon$ where the latent tensor $\tensor{X}$ can be represented exactly by a Tucker model with a low multilinear rank and $\varepsilon$ denotes i.i.d. Gaussian noise.

Given an incomplete tensor, Bayesian Tucker model only considers the observed entries, yielding a new likelihood function represented by
\begin{equation}
\nonumber
\begin{split}
p\big(\tensor{Y}_\Omega\vert\{\mathbf{U}^{(n)}\},\tensor{G},\tau\big) &=\prod_{(i_1,\ldots, i_N)\in\Omega} \mathcal{N}\left(\mathcal{Y}_{i_1\ldots i_N} \vert \mathcal{X}_{i_1\ldots i_N}, \tau^{-1}\right).
\end{split}
\end{equation}
Based on Tucker decomposition framework, we can thus represent the observation model as that $\forall (i_1,\ldots, i_N)$,
{\small
\begin{equation}
\label{eq:BTDClikelihood}
\begin{split}
\mathcal{Y}_{i_1\cdots i_N } \Big| \left\{\mathbf{u}^{(n)}_{i_n}\right\}, \tensor{G}, \tau   & \sim  \mathcal{N}\left(\left(\bigotimes_n \mathbf{u}_{i_n}^{(n)T}\right)\text{vec}(\tensor{G}), \,\tau^{-1}\right)^{\mathcal{O}_{i_1\cdots i_N}}. \\
\end{split}
\end{equation}
}
For Tucker decomposition of an incomplete tensor, the problem is ill-conditioned and has infinite solutions. The low-rank assumption play an key role for successful tensor completion, which implies that the determination of multilinear rank significantly affects the predictive performance. However, standard model selection strategies, such as cross-validation, cannot be applied for finding the optimal multilinear rank because it varies dramatically with missing ratios. Therefore, the inference of multilinear rank is more challenging when missing values occur.

As described in BTD model, we employ two types of hierarchical group sparsity priors over the factor matrices and core tensor with aim to seek the minimum multilinear rank automatically, which is more efficient and elegant than the standard model selections by repeating many times and selecting one optimum model. Therefore, the model priors for all hidden variables are same with that in BTD model. By combining likelihood model in (\ref{eq:BTDClikelihood}) with the model priors in (\ref{eq:BTDprior}), we propose a Bayesian Tucker Completion (BTC) model, which enable us to infer the minimum multilinear rank as well as the noise level solely from partially observed data without requiring any tuning parameters.

\subsection{Model inference}
For BTC model, we also employ VB inference framework to learn the model under a fully Bayesian treatment. Since BTC model differs from BTD model in the likelihood function (\ref{eq:BTDClikelihood}), indicating that the inference for factor matrices, core tensor and noise parameter are essentially different, while other hyperparameters can be inferred by the same solutions. In this section, we present only the main solutions while the detailed derivations are provided in the Appendix.

As can be derived, the variational posterior distribution over the core tensor $\tensor{G}$ is
\begin{equation}
q(\tensor{G}) = \mathcal{N}\left(\text{vec}(\tensor{G})\big\vert \text{vec}(\widetilde{\tensor{G}}), \Sigma_G \right)
\end{equation}
where the posterior parameters can be updated by
{\small
\begin{gather}
\label{eq:InferCoreMean}
\text{vec}(\widetilde{\tensor{G}}) = \mathbb{E}[\tau]\, \Sigma_G \!\!\!\!\!\!\sum_{(i_1,\ldots, i_N)\in\Omega}\left( \mathcal{Y}_{i_1\cdots i_N} \bigotimes_n \mathbb{E}\left[\mathbf{u}^{(n)}_{i_n}\right]\right) \\
\label{eq:InferCoreVar}
\Sigma_G =  \left\{\mathbb{E}[\beta]\bigotimes_n\mathbb{E}\left[\boldsymbol\Lambda^{(n)}\right] + \mathbb{E}[\tau]\!\!\!\!\!\! \sum_{(i_1,\ldots,i_N)\in\Omega}\!\!\bigotimes_n\mathbb{E}\left[\mathbf{u}^{(n)}_{i_n}\mathbf{u}^{(n)T}_{i_n}\right] \right\}^{-1}
\end{gather}
} It should be noted that Theorems \ref{theorem:1}, \ref{theorem:2} and multilinear operations cannot be applied to (\ref{eq:InferCoreVar}) due to the sum of \emph{kronecker} products. Thus the sequential \emph{kronecker} products must be computed explicitly, resulting in the computational cost of $\mathcal{O}\left(\prod_n R_n^3 + M\prod_n R_n^2\right)$, where $M$ denotes the number of observed entries (i.e., data size), and memory cost of $\mathcal{O}\left(\prod_n R_n^2\right)$. This severely prevents the method from being applied to large-scale datasets. To improve scalability, we propose an alternative solution by optimizing $\arg min_{\tensor{G}}\{ -\ln q(\tensor{G})\}$ instead of closed-form update in (\ref{eq:InferCoreMean}). This can be achieved by employing a nonlinear conjugate gradient method with the gradient given by
\begin{multline}
\frac{\partial \left\{-\ln q(\tensor{G})\right\}}{\partial(\text{vec}(\tensor{G}))}  = \mathbb{E}[\beta]\left(\bigotimes_n\mathbb{E}\left[\boldsymbol\Lambda^{(n)}\right]\right)\text{vec}(\tensor{G})  \\
+\mathbb{E}[\tau]\!\!\!\!\!\!\sum_{(i_1,\ldots,i_N)\in\Omega}\left\{\left(\bigotimes_n\mathbb{E}\left[\mathbf{u}^{(n)}_{i_n}\mathbf{u}^{(n)T}_{i_n}\right]\right)\text{vec}(\tensor{G})\right\}\\   -\mathbb{E}[\tau]\!\!\!\!\!\! \sum_{(i_1,\ldots, i_N)\in\Omega}\left( \mathcal{Y}_{i_1\cdots i_N} \bigotimes_n \mathbb{E}\left[\mathbf{u}^{(n)}_{i_n}\right]\right)
\end{multline}
Thus, multilinear operations can be applied without explicitly computation of kronecker products, resulting in reduced computational complexity of $\mathcal{O}\left(M\prod_n R_n\right)$ and reduced memory cost of $\mathcal{O}\left(\prod_n R_n\right)$, which scales linearly with data size and model complexity.

As can be derived, the variational posterior distribution over $\big\{\mathbf{U}^{(n)}\big\}$ is factorized as
\begin{equation}
\label{eq:PosterFactorU}
q\big(\mathbf{U}^{(n)}\big) = \prod_{i_n} \mathcal{N}\left(\mathbf{u}_{i_n}^{(n)}\middle\vert \widetilde{\mathbf{\mathbf{u}}}^{(n)}_{i_n}, \Psi^{(n)}_{i_n} \right),\quad  n=1,\ldots,N,
\end{equation}
where the posterior parameters can be updated by
{\small
\begin{gather}
\label{eq:InferFactorUmean}
\widetilde{\mathbf{\mathbf{u}}}^{(n)}_{i_n} = \mathbb{E}[\tau] \Psi^{(n)}_{i_n}\mathbb{E}\left[\mathbf{G}_{(n)}\right]\!\!\!\!\!\!\sum_{(i_1,\ldots, i_N)\in\Omega}\left(\mathcal{Y}_{i_1\cdots i_N}\bigotimes_{k\neq n}\mathbb{E}\left[\mathbf{u}^{(k)}_{i_k}\right] \right) \\
\label{eq:InferFactorUvar}
\Psi^{(n)}_{i_n} = \left\{\mathbb{E}[\boldsymbol\Lambda^{(n)}] + \mathbb{E}[\tau]\mathbb{E}\left[\mathbf{G}_{(n)} \boldsymbol\Phi^{(n)}_{i_n}\mathbf{G}_{(n)}^T\right]\right\}^{-1}
\end{gather}
}
where $\boldsymbol\Phi^{(n)}_{i_n} =\!\!\!\!\!\displaystyle\sum_{(i_1,\ldots, i_N)\in\Omega} \, \bigotimes_{k\neq n}\mathbf{u}^{(k)}_{i_k}\mathbf{u}^{(k)T}_{i_k}$, the summation is performed over the observed data locations whose \mbox{mode-$n$} index is fixed to $i_n$. In other words, $\boldsymbol\Phi^{(n)}_{i_n}$ represents the statistical information of mode-$k$ ($k\neq n$) latent factors that interact with $\mathbf{u}^{(n)}_{i_n}$. In (\ref{eq:InferFactorUvar}), the complex posterior expectation can be computed by
{\small
\begin{multline}
\!\!\!\!\!\text{vec}\left\{\mathbb{E}\left[\mathbf{G}_{(n)}\boldsymbol\Phi^{(n)}_{i_n}\mathbf{G}_{(n)}^T\right]\right\}=
\mathbb{E}\left[ \mathbf{G}_{(n)}\otimes \mathbf{G}_{(n)} \right]\text{vec}\left(\boldsymbol\Phi^{(n)}_{i_n}\right).
\end{multline}
}The computational complexity for inference of $\mathbf{u}_{i_n}^{(n)}$ is $\mathcal{O}\left(R^3_n + M\prod_n R_n \right)$, while the memory cost is $\mathcal{O}(\prod_n R_n)$.

An intuitive interpretation of (\ref{eq:InferFactorUvar}) is that the posterior covariance $\Psi^{(n)}_{i_n}$ combines prior information denoted by $\mathbb{E}[\boldsymbol\lambda^{(n)}]$ and posterior information of interacted factors in other modes, while the tradeoff between these two terms is controlled by model residual $\mathbb{E}[\tau]$. Hence, if updated $\mathbb{E}[\boldsymbol\lambda^{(n)}_{r_n}]$ is quite large and model is not well fitted, then the posterior variance of $r_n$th component will be easily forced to zero.

Given updated $q\big(\mathbf{u}_{i_n}^{(n)}\big)$ from (\ref{eq:PosterFactorU}), we can easily compute the posterior expectations of some expressions as
\begin{equation}
\begin{split}
\mathbb{E}\big[\mathbf{u}^{(n)T}_{\cdot r_n}\mathbf{u}^{(n)}_{\cdot r_n} \big] &= \widetilde{\mathbf{u}}_{\cdot r_n}^{(n)T}\widetilde{\mathbf{u}}_{\cdot r_n}^{(n)} + \sum_{i_n} \big(\Psi^{(n)}_{i_n}\big)_{r_n r_n},\\
\mathbb{E}\big[\mathbf{u}^{(n)}_{i_n}\mathbf{u}^{(n)T}_{i_n}\big] &= \widetilde{\mathbf{u}}_{i_n}^{(n)}\widetilde{\mathbf{u}}_{i_n}^{(n)T} + \Psi^{(n)}_{i_n}.
\end{split}
\end{equation}

The variational posterior distribution over noise precision $\tau$ can be derived as $q(\tau) = Ga(\tau\vert a^\tau_M,b^\tau_M)$ whose parameters are updated by
{\small
\begin{equation}
\begin{split}
a^\tau_M &= a^\tau_0 +  \frac{1}{2}\sum_{i_1,\ldots,i_N} O_{i_1\cdots i_N},\\
b^\tau_M &= b_0^\tau+\frac{1}{2} \mathbb{E}\left[\sum_{(i_1,\ldots,i_N)\in\Omega}\!\!\!\left(\mathcal{Y}_{i_1\cdots i_N}-\left(\bigotimes_n\mathbf{u}^{(n)T}_{i_n}\right)\text{vec}(\tensor{G})\right)^2\right]
\end{split}
\end{equation}
}The posterior expectation of model residuals over observed entries can be evaluated by
{\small
\begin{multline}
\mathbb{E}\left[\sum_{(i_1,\ldots,i_N)\in\Omega}\!\!\!\left(\mathcal{Y}_{i_1\cdots i_N}-\left(\bigotimes_n\mathbf{u}^{(n)T}_{i_n}\right)\text{vec}(\tensor{G})\right)^2\right]=\\
 \|\tensor{Y}_{\Omega}\|_F^2 - 2\left(\sum_{(i_1,\ldots,i_N)\in\Omega} \!\!\!\!\!\!\mathcal{Y}_{i_1\cdots i_N}\bigotimes_n \mathbb{E}\left[\mathbf{u}^{(n)T}_{i_n}\right]\right)\mathbb{E}[\text{vec}(\tensor{G})] \\
  +\text{Tr}\left\{\mathbb{E}\left[\text{vec}(\tensor{G})\text{vec}(\tensor{G})^T \right]\!\!\!\!\!\sum_{(i_1,\ldots,i_N)\in\Omega}  \bigotimes_n \mathbb{E}\left[ \mathbf{u}_{i_n}^{(n)}\mathbf{u}_{i_n}^{(n)T}\right]  \right\}.
\end{multline}
} The computational complexity is $\mathcal{O}(M\prod_n R_n)$, if multilinear operations have been applied.

Note that other hidden hyperparameters in $\Theta$ except $\{\tensor{G}, \{\mathbf{U}^{(n)}\}, \tau\}$ can be inferred essentially by the same solutions with BTD model. In addition, the lower bound $\mathcal{L}(q) = \mathbb{E}_{q(\Theta)}[\ln p(\tensor{Y}_\Omega,\Theta)] + H(q(\Theta))$ can be also evaluated with different expressions related to $\{\tensor{G}, \{\mathbf{U}^{(n)}\}, \tau\}$ (see Appendix for details).

The predictive distributions over missing entries, given observed entries, can be approximated by using variational posterior distributions $q(\Theta)$ as follows
\begin{equation}
\begin{split}
&p(\mathcal{Y}_{i_1\cdots i_N} \vert \tensor{Y}_{\Omega}) = \int p(\mathcal{Y}_{i_1\cdots i_N} \vert \Theta)p(\Theta\vert \tensor{Y}_{\Omega}) \, \text{d}\Theta\\
&\approx\mathcal{N} \left(\mathcal{Y}_{i_1\cdots i_N} \middle| \tilde{\mathcal{Y}}_{i_1\cdots i_N}, \; \mathbb{E}[\tau]^{-1} + \sigma^2_{i_1\cdots i_N} \right),
\end{split}
\end{equation}
where the posterior parameters can be obtained by
{\small
\begin{equation}
\begin{split}
\tilde{\mathcal{Y}}_{i_1\cdots i_N} &=
 \left(\bigotimes_n \mathbb{E}\left[\mathbf{u}^{(n)T}_{i_n}\right]\right) \mathbb{E}[\text{vec}(\tensor{G})]\\
 \sigma^2_{i_1\cdots i_N} &= \text{Tr}\left(\mathbb{E}\left[\text{vec}(\tensor{G})\text{vec}(\tensor{G})^T \right] \bigotimes_n \mathbb{E}\left[ \mathbf{u}_{i_n}^{(n)}\mathbf{u}_{i_n}^{(n)T}\right]\right) \\
 - \mathbb{E}&[\text{vec}(\tensor{G})]^T \left(\bigotimes_n \mathbb{E}\left[\mathbf{u}^{(n)}_{i_n}\right]\mathbb{E}\left[\mathbf{u}^{(n)T}_{i_n}\right]\right) \mathbb{E}[\text{vec}(\tensor{G})].
\end{split}
\end{equation}
}Therefore, our model can provide not only predictions over missing entries, but also the uncertainty of predictions, which is quite important for some specific applications.

\section{Algorithm Related Issues}
\label{sec:algorithm}
We denote Bayesian sparse Tucker models (i.e., BTD and BTC) when two different group sparsity priors have been employed  by BTD-T, BTD-L, BTC-T, BTC-L respectively, where T, L represent a hierarchical Student-t and Laplace priors. The main procedure of model inference is described as follows:
\begin{itemize}
  \item \textbf{Setting of top-level hyperparameters.} For BTD-T and BTC-T models, $\{a_0^\tau,b_0^\tau, a_0^\beta, b_0^\beta, a_0^\lambda,b_0^\lambda\}$ are set to 1e-9. For BTD-L and BTC-L models, $\{a_0^\tau,b_0^\tau, a_0^\beta, b_0^\beta, a_0^\gamma,b_0^\gamma\}$ are set to 1e-9. This setting results in a noninformative prior, which ensures that model inference is solely based on observed data.
  \item \textbf{Initialization of hidden parameters $\Theta$.} For BTD-T and BTC-T, $\Theta =\{\tensor{G},\{\mathbf{U}^{(n)}\}, \{\boldsymbol\lambda^{(n)}\}, \tau, \beta\}$. For BTD-L and BTC-L, $\Theta =\{\tensor{G},\{\mathbf{U}^{(n)}\}, \{\boldsymbol\lambda^{(n)}\}, \tau, \beta, \gamma\}$. $\tensor{G}$ does not need to be initialized due to that it will be updated firstly. The latent dimensions can be initialized by $R_n=I_n, \forall n$ or manually by $R_n<I_n$. $\{\mathbf{U}^{(n)}\}$ can be initialized as left singular vectors by mode-n SVD, while an alternative one is sampling from $\mathcal{N}(0,1)$. $\tau$ is initialized by $1/\sigma^2_{\tensor{Y}}$ denoting an inverse of data variance. $\{\boldsymbol\lambda^{(n)}\}, \beta, \gamma$ are simply initialized to be 1.
  \item \textbf{Variational model inference.} For all the models, the approximate inference  can be updated sequentially in the order of $\{\tensor{G},\{\mathbf{U}^{(n)}\}, \beta, \{\boldsymbol\lambda^{(n)}\},  \gamma, \tau\}$.
  \item \textbf{Model reduction.}  Pruning out zero-valued latent components from $\{\mathbf{U}^{(n)}\}$ as well as the associated slices from $\tensor{G}$ and associated components from $\{\boldsymbol\lambda^{(n)}\}$.
  \item \textbf{Lower bound of model evidence.} The lower bound of log-marginal likelihood is evaluated according to specific models, which can be used to test for convergence.
  \item \textbf{Predictive distribution.} For BTD models, the predictive distribution can be inferred for recovering latent tensor without noises. For BTC modes, it can be inferred for recovering missing entries.
\end{itemize}

For BTD-T and BTD-L models, the overall computational complexity is $\mathcal{O}(\sum_n R_n^3  + N\prod_n I_n)$, while the memory cost is $\mathcal{O}(\prod_n R_n + \prod_n I_n)$. For BTC-T and BTC-L models, the overall computational complexity is $\mathcal{O}(\sum_n I_n R_n^3 + M \prod_n R_n \sum_n I_n)$, while the memory cost is $\mathcal{O}(\prod_n R_n + \sum_n I_nR_n)$. Therefore, BTD models are more computational efficient than BTC models, but it requires more memory. The computational complexity of all models scales linearly with data size, but polynomially with model complexity. Hence, our models are suitable for relatively low multilinear rank tensors. It should be noted that, because of automatic model reduction, $\{R_1,\ldots, R_N\}$ reduces rapidly in the first few iterations, resulting in that computational complexity will be decreased with number of iterations.

The key advantage of our models is automatic model determination (i.e., learning multilinear rank), which enable us to obtain an optimal low-rank Tucker approximation from a noisy and incomplete tensor. Secondly, taking into account uncertainty information of all model parameters by full posterior inference, our method can effectively prevent overfitting problem and provide predictive uncertainty as well. Thirdly, a deterministic Bayesian inference with closed-form update rules is derived for model learning, which is more efficient and scalable than sampling based inference. Finally, our methods are significantly convenient for practical applications since they do not require any tuning parameters.

\section{Experimental Results}
\label{sec:experiment}
We evaluated the proposed models and algorithms by extensive experiments and comparisons with state-of-the-art related works which include HOOI~\cite{kolda2009tensor}, ARD-Tucker~\cite{morup2009automatic}, iHOOI~\cite{liu2014generalized}, WTucker~\cite{filipovic2013tucker}, HaLRTC~\cite{liu2013tensorcompletion}.
In all our experiments, the parameters of HOOI, WTucker, and HaLRTC were tuned carefully and the best possible performances were reported. ARD-Tucker and iHOOI can automatically adapt tensor rank, which is similar to BTD and BTC methods proposed in this paper.

\subsection{Synthetic data}
The simulated data was generated according to an orthogonal Tucker model. More specifically, the core tensor was drawn from $\mathcal{G}_{r_1r_2r_3}\sim\mathcal{N}(0,1), \forall r_n=1,\ldots, R_n$. The factor matrices were drawn from $\mathbf{u}_{i_n r_n}^{(n)}\sim\mathcal{N}(0,1), \forall i_n=1,\ldots,50, \forall r_n=1,\ldots R_n, \forall n=1,2,3$, and then orthogonalization was performed such that $\mathbf{U}^{(n)T}\mathbf{U}^{(n)} =\mathbf{I}_{R_n}$. In principle, such data can be perfectly fitted by HOSVD or a more powerful method HOOI.

\begin{figure}[h]
\centering
\subfigure[\footnotesize HOOI]{
   \includegraphics[width=0.4\columnwidth] {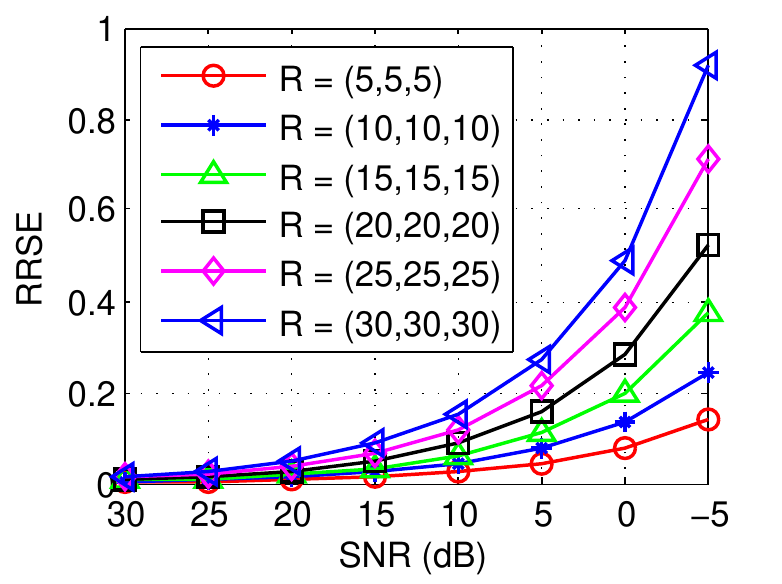}
   \label{fig:sim1}
 }
 \subfigure[\footnotesize BTD-T]{
   \includegraphics[width=0.4\columnwidth] {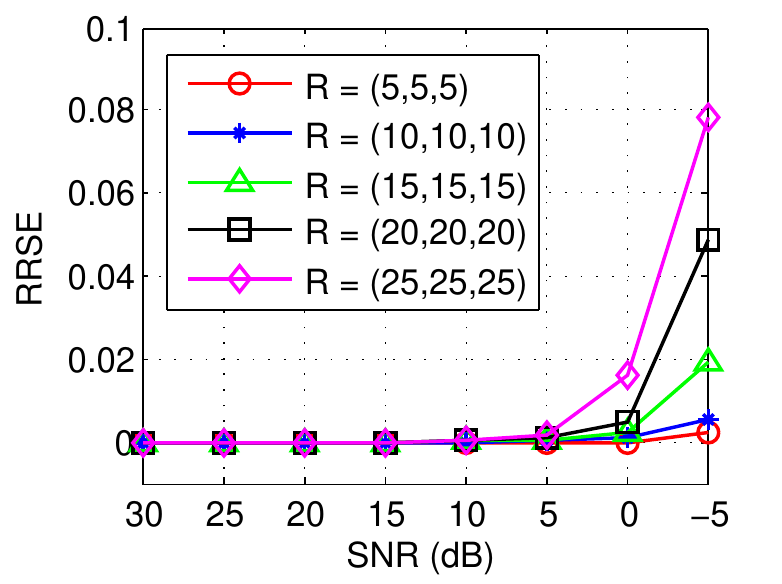}
   \label{fig:sim2}
   }
    \subfigure[\footnotesize BTD-L]{
   \includegraphics[width=0.4\columnwidth] {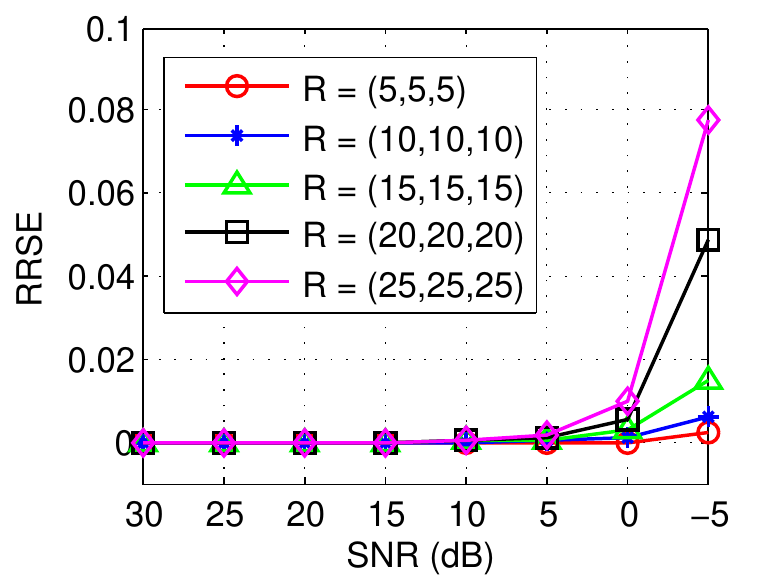}
   \label{fig:sim3}
   }
    \subfigure[\footnotesize ARD-Tucker]{
   \includegraphics[width=0.4\columnwidth] {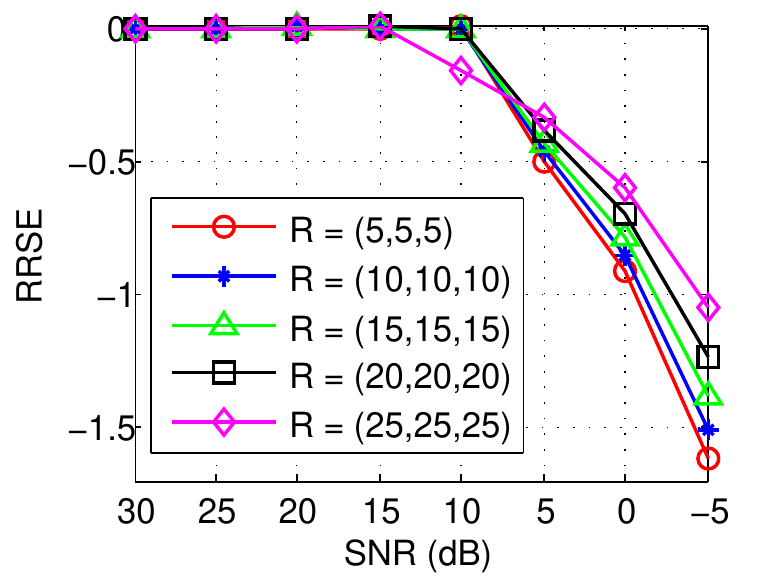}
   \label{fig:sim4}
   }
\caption{\subref{fig:sim1} shows performance baseline using true tensor rank under varying noise levels (SNR) and tensor rank (R). \subref{fig:sim2}\subref{fig:sim3}\subref{fig:sim4} show discrepancy between baseline and three methods which can infer rank automatically. }
\label{fig:RRSESim1}
\end{figure}

\begin{figure}
  \centering
  \includegraphics[width=0.7\columnwidth,height=0.7in]{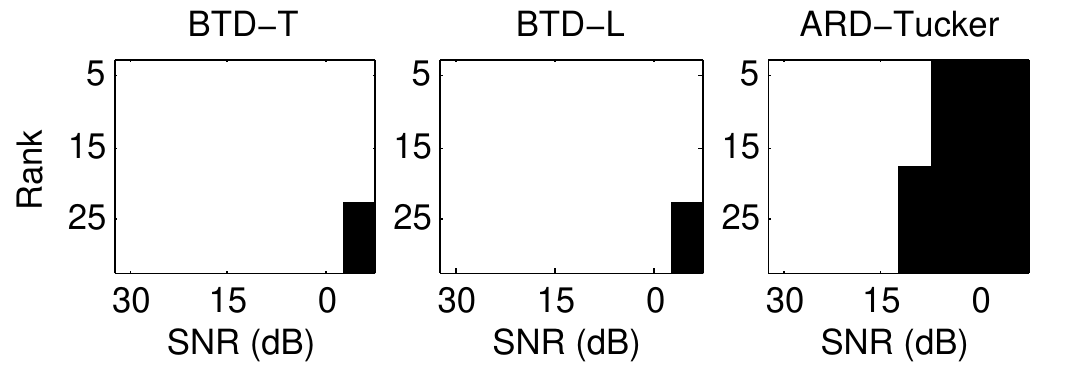}\\
  \caption{Accuracy of inferred tensor rank under conditions of varying SNR and R. White area indicates successful estimations and black area indicates failed cases.}
  \label{fig:ErrRankSim1}
\end{figure}

\begin{figure}
  \centering
  \includegraphics[width=0.8\columnwidth,height=0.8in]{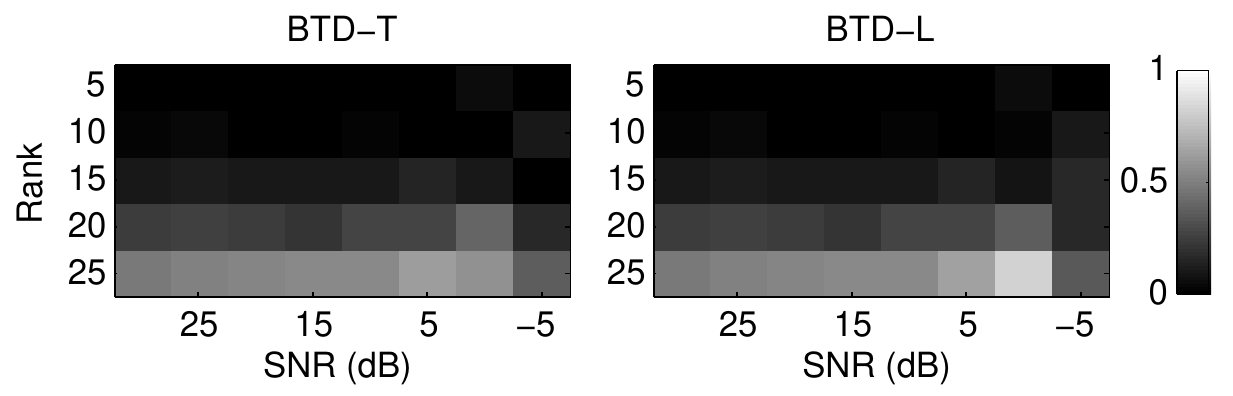}\\
  \caption{Absolute error of inferred SNR. The maximum error is close to 1dB when R=(25, 25, 25) and SNR= -5dB.}
  \label{fig:ErrSNRSim1}
\end{figure}

We evaluated BTD-T/L on complete tensors under varying noise levels with SNR ranging from 30dB to -5dB and true tensor rank ($R_1,R_2,R_3$) ranging from 5 to 25. The reconstruction performance was evaluated by RRSE =$\|\tilde{\tensor{X}}-\tensor{X}\|_F/\|\tensor{X}\|_F$ where $\tilde{\tensor{X}}$ denotes the estimation of true latent tensor $\tensor{X}$. Theoretically, the best performance should be obtained by HOOI given true tensor rank. As shown in Fig.~\ref{fig:RRSESim1}, RRSE of HOOI increases exponentially with noise levels and larger $R$ leads to a more rapid degeneration of performance. It should be emphasized that BTD-T/L achieve surprisingly improved performances without knowing true tensor rank, and the improvements become more significant for larger $R$ and smaller SNR. The RRSEs obtained by BTD-T and BTD-L are almost equal under all conditions. ARD-Tucker, which can also automatically infer tensor rank, however obtained significantly degenerated performance when SNR$<$10dB. The accuracy of rank determination is compared in Fig.~\ref{fig:ErrRankSim1}, showing that BTD-T/L are able to exactly infer the tensor rank only except the cases of SNR=-5dB, R$>=$25 and outperform ARD-Tucker. In addition, BTD-T/L can infer noise levels accurately as shown in Fig.~\ref{fig:ErrSNRSim1}. The averaged runtime of BTD-T/L are 4/3 seconds while ARD-Tucker needs 47 seconds.


\begin{figure}[h]
\centering
\subfigure[\footnotesize BTC-T]{
   \includegraphics[width=0.4\columnwidth] {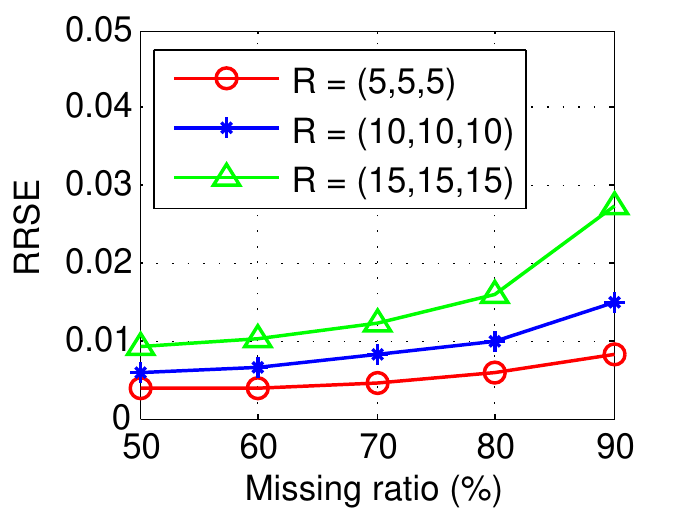}
   \label{fig:sim1}
 }
 \subfigure[\footnotesize BTC-L]{
   \includegraphics[width=0.4\columnwidth] {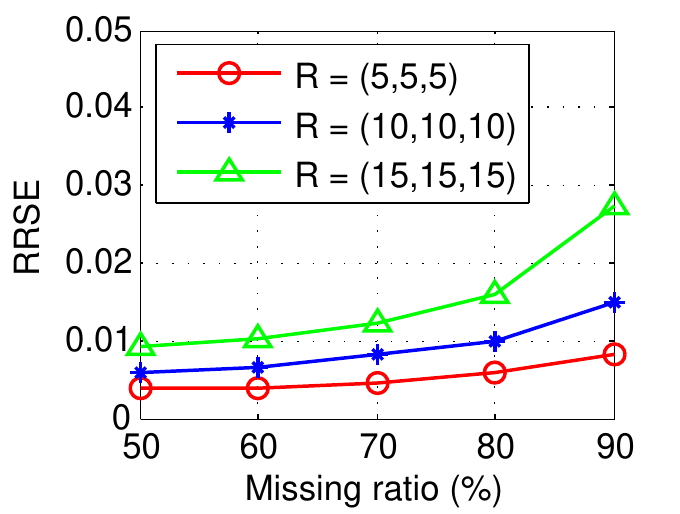}
   \label{fig:sim2}
   }
    \subfigure[\footnotesize iHOOI]{
   \includegraphics[width=0.4\columnwidth,height=0.9in] {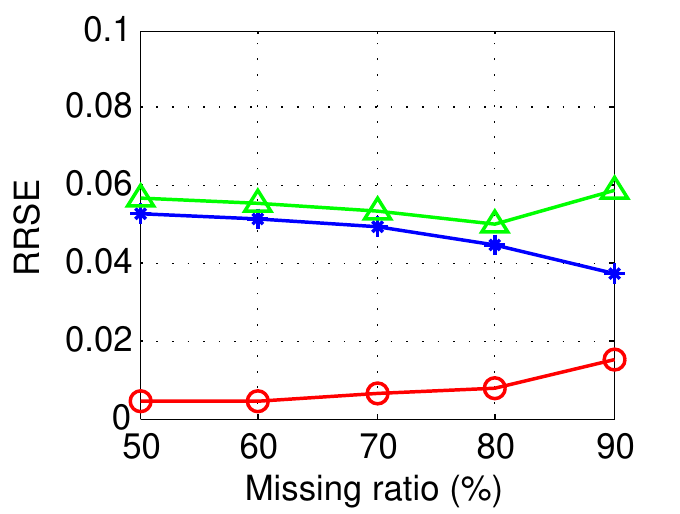}
   \label{fig:sim3}
   }
    \subfigure[\footnotesize HaLRTC]{
   \includegraphics[width=0.4\columnwidth,height=0.9in] {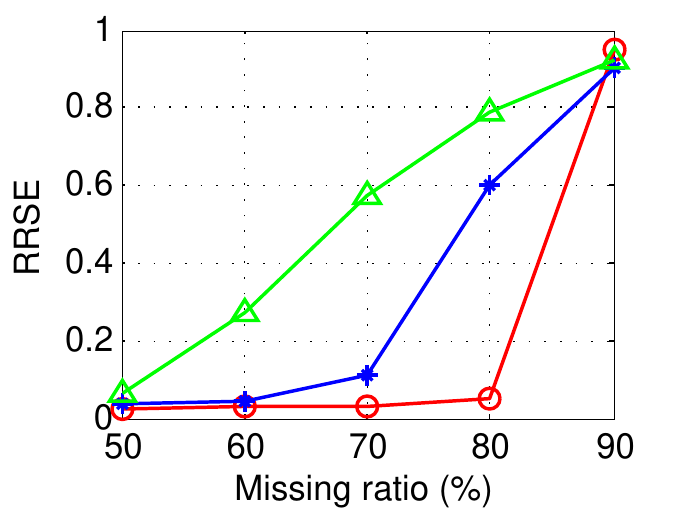}
   \label{fig:sim4}
   }
\caption{Tensor completion performances obtained by different methods under varying conditions of MR and true rank. The legend for all figures are same. }
\label{fig:RRSESim2}
\end{figure}

\begin{figure}[h]
  \centering
  \includegraphics[width=0.8\columnwidth]{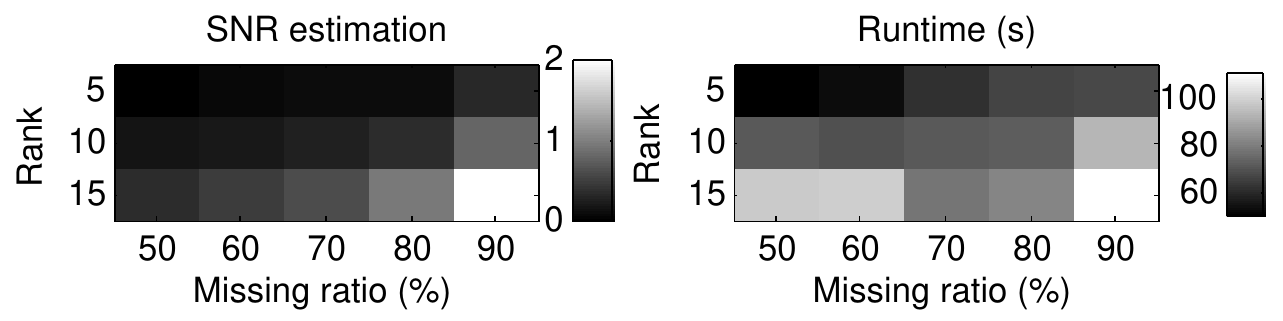}\\
  \caption{Absolute error of inferred SNR whose maximum is 2dB when R = (15,15,15) and MR = 90\%. The runtime is proportional to the true rank. }
  \label{fig:ErrSNRSim2}
\end{figure}

We evaluated BTC-T/L on incomplete tensors under varying missing ratios (MR) ranging from 50\% to 90\% and true rank $\{R_n\}$ ranging from 5 to 15, while SNR was fixed to 30dB. As shown in Fig.~\ref{fig:RRSESim2}, BTC-T/L perform similarly under all conditions and significantly outperform iHOOI and HaLRTC. The RRSE increases with MR and larger R leads to a more rapid increase. iHOOI and HaLRTC, which are able to optimize tensor rank base on nuclear norm, perform well on relatively low rank tensors but degenerates rapidly with increasing rank. It should be emphasized that BTC-T/L exactly inferred tensor rank with 100\% accuracy under all these conditions. As shown in Fig.~\ref{fig:ErrSNRSim2}, BTC-T/L can also accurately infer SNR when either true rank or MR is low, and the runtime depends on true rank. In summary, if both MR and true rank increase, the completion performance will decrease.

\subsection{Chemometrics data}
We evaluated BTD and BTC models on chemometrics data analysis\footnote{Datasets are available from repository for multi-way data analysis. \url{http://www.models.kvl.dk/datasets}} in terms of recovering the number of latent components which have meaningful interpretation and imputing missing data. Five tested data sets are Amino Acids Fluorescence ($5\times 201\times 61$), Flow Injection ($12\times 100 \times 89$), Eletronic Nose ($18\times 241\times 12$), Sugar Process ($268\times 571\times 7$), and Enzymatic Acivity ($3\times 3 \times 3 \times 3 \times 5$). For Amino Acids Fluorescence data, each sample contains different amounts of tyrosine, tryptophan and phenylalanine dissolved in phosphate buffered water and measured by fluorescence on a PE LS50B spectrofluorometer. Although each sample is represented as a full rank matrix, leading to a full rank tensor.  Ideally it should be describable by three latent components corresponding to individual amino acids.

We tested BTD-T/L methods on the original data as well as noise corrupted data (SNR=0dB). Since the standard Tucker decomposition (HOOI) needs a predefined tensor rank, it can be specified as the inferred rank from BTD-T. As shown in Table~\ref{tab:chemometrics1}, tensor rank inferred by BTD-T is $(3,3,3)$, indicating that the underlying components associated to three amino acids are successfully captured. In addition, BTD-T yields a stable rank estimation even for highly noisy data. The performances of BTD-T/L are similar and even better than HOOI in most cases, although the same rank was employed. As compared to ARD-Tucker, BTD-T/L always obtain better performance and are more robust to noise. The computation efficiency of BTD-T/L is also significantly higher than ARD-Tucker.

We also tested BTC-T/L methods by randomly selecting 90\% data for training, and the remainder was used as test case. We repeated this partition 10 times. Table~\ref{tab:chemometrics2} presents predictive performances and stability of inferred rank. Note that HaLRTC is sensitive to tuning  parameters, and the best possible performances by searching tuning parameters in a wide range are reported. iHOOI is able to learn tensor rank automatically, however, the inferred rank is unstable and sensitive to different partitions. In contrast, BTC-T/L achieved the best predictive performance on all five datasets, while the inferred rank is stable under any specific missing ratio.

\begin{table*}[!htb]
\renewcommand{\arraystretch}{1.2}
\caption{\small Chemometrics data analysis by tensor decompositions. The SNR of noisy data is 0dB. $R$ denotes inferred tensor rank. N/A indicates an inapplicable case. Runtime is measured by seconds. }
\label{tab:chemometrics1}
\centering
\resizebox{1\textwidth}{!}
{
\begin{tabular}{ c | c |c | c | c | c | c | c | c | c | c | c}
\hline
 & & \multicolumn{2}{c|}{Amino}  & \multicolumn{2}{c|}{Flow} & \multicolumn{2}{c|}{Electronic}  & \multicolumn{2}{c|}{Sugar} & \multicolumn{2}{c}{Enzymatic}  \\
 \hline
 & & Original & Noisy & Original & Noisy & Original & Noisy & Original & Noisy & Original & Noisy\\
\hline
 HOOI   & RRSE & 0.0245 &0.0960 & 0.0321 & 0.0827 & 0.0328 & 0.0664 & 0.0386 & 0.0839 &0.1677 & 0.2174\\
\hline
\multirow{3}{*}{ARD-Tucker} & R & (4, 3, 3) &(6, 3, 3) & (5, 4, 3) & (12, 48, 49) & (1, 1, 1) & (3, 2, 3) & (5, 8, 7) & (5, 8, 7) & N/A & N/A\\
                       & RRSE & 0.0260 &0.0961 & 0.0476 & 0.4377 & 0.0328 & 0.1006 & 0.0387 & 0.0788 & N/A & N/A\\
                        & Runtime & 32 & 38 & 29 & 275 & 22 & 49 & 138 & 145 & N/A & N/A\\
  \hline
\multirow{2}{*}{BTD-T} & R    & (3, 3, 3) &(3, 3, 3) & (3, 5, 3) & (3, 4, 3) & (1, 2, 1) & (1, 46, 1) & (6, 24, 6) & (5, 11, 6) &(1,1,1,1,1) & (1,1,1,1,1)\\
                       & RRSE & 0.0245 &0.0950 & 0.0321 & 0.0812 & 0.0328 & 0.0664 & 0.0387 & 0.0788 &0.1677 & 0.2216\\
                       & Runtime & 2 & 2 & 2 & 2 & 2 & 5 & 15 & 15 & 3 & 8\\
\hline
\multirow{2}{*}{BTD-L} & R    & (3, 6, 3) &(3, 3, 3) & (3, 5, 3) & (3, 4, 3) & (1, 1, 1) & (1, 24, 1) & (7, 25, 6) & (5, 11, 6) &(1,1,1,1,1) & (1,1,1,1,2)\\
                       & RRSE & 0.0222 &0.0950 & 0.0321 & 0.0812 & 0.0328 & 0.0664 & 0.0353 & 0.0789 &0.1677 & 0.2209\\
                       & Runtime & 5 & 4 & 2 & 4 & 2 & 5 & 15 & 15 & 8 & 7\\
\hline
\end{tabular}
}
\end{table*}

\begin{table*}[!htb]
\renewcommand{\arraystretch}{1.2}
\caption{\small Chemometrics data completion with 90\% missing ratio. Std(R) denotes standard deviation of inferred tensor rank.  }
\label{tab:chemometrics2}
\centering
\resizebox{1\textwidth}{!}
{
\begin{tabular}{  c |c | c | c | c | c | c | c | c | c | c}
\hline
 & \multicolumn{2}{c|}{Amino}  & \multicolumn{2}{c|}{Flow} & \multicolumn{2}{c|}{Electronic}  & \multicolumn{2}{c|}{Sugar} & \multicolumn{2}{c}{Enzymatic}  \\
 \hline
 & Std(R) & RRSE & Std(R) & RRSE & Std(R) & RRSE & Std(R) & RRSE & Std(R) & RRSE\\
\hline
HaLRTC  & N/A & 0.35$\pm$0.01 & N/A & 0.12$\pm$ 0.02 & N/A & 0.06$\pm$0.01 & N/A & 0.18$\pm$0.00 & N/A & 0.68$\pm$0.04\\
\hline
iHOOI   & (0,51,17) &0.64$\pm$0.08 & (0.3,0.3,0.3)  & 0.44$\pm$0.32 & (5,38,2) & 0.04$\pm$0.01 & (0,2,0) & 0.64$\pm$0.00 &(0,0,0,0.3,0.5) & 0.89$\pm$0.49\\
\hline
BTC-T   & (0.4,0.4,0) &0.03$\pm$0.00 & (0.5,0.6,0.4) & 0.10$\pm$0.04 & (0,0,0) & 0.03$\pm$0 & (2,1.5,0) & 0.07$\pm$0.01 & (0,0,0,0,0) & 0.21$\pm$0.03\\
\hline
BTC-L   & (0.5,0.4,0) &0.026$\pm$0.00 & (0.5,0.5,0.3) & 0.11$\pm$0.04 & (0,0,0) & 0.04$\pm$0 & (1.4,0.7,0) & 0.07$\pm$0.01 & (0.4,0,0,0,0) & 0.21$\pm$0.03\\
\hline
\end{tabular}
}
\end{table*}

\subsection{MRI data}
\begin{figure}[h]
\centering
\subfigure[50\% missing]{
   \includegraphics[width=0.47\columnwidth] {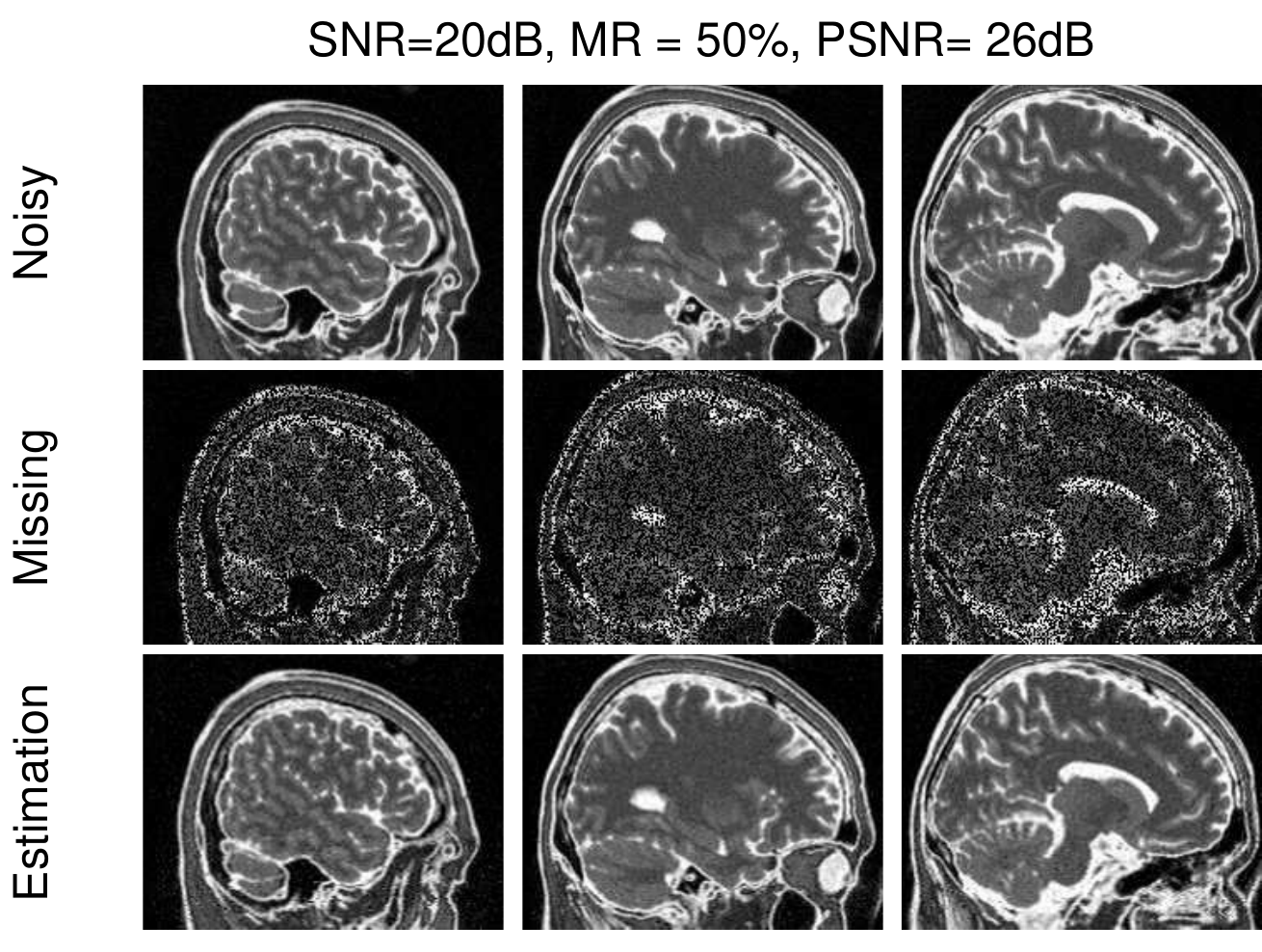}
   \label{fig:sim1}
 }
 \subfigure[80\% missing]{
   \includegraphics[width=0.47\columnwidth] {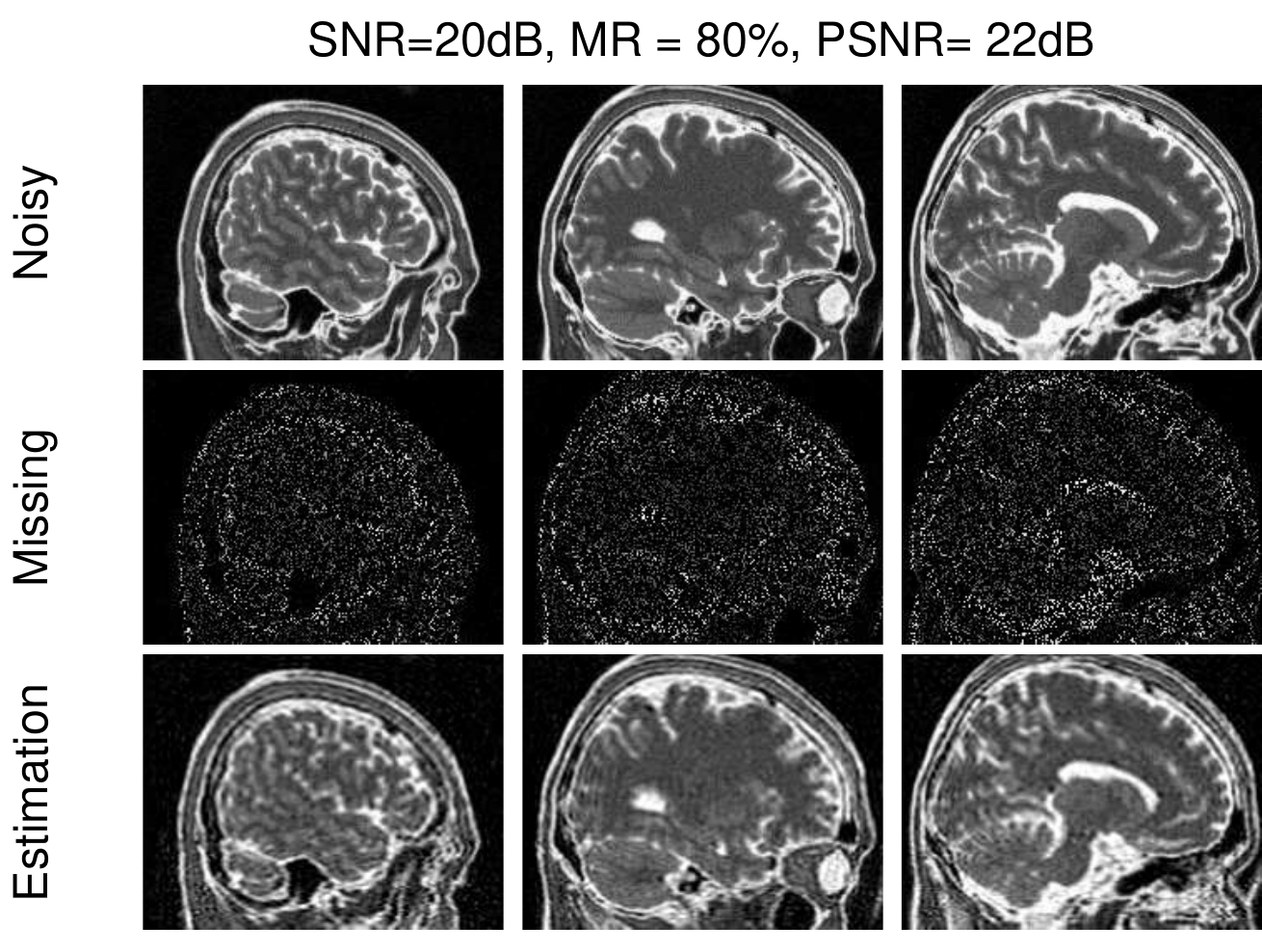}
   \label{fig:sim2}
   }
\caption{Visualization of MRI data completion obtained by BTC-T. }
\label{fig:MRIvisual}
\end{figure}

In this section, we evaluate BTC-T/L methods by MRI data~\footnote{Dataset is available from \url{http://brainweb.bic.mni.mcgill.ca/brainweb/}}, particularly for the generalization performance when original data does not posses a global low-rank structure. Since the dimensions and rank of MRI data are generally high, predictions of missing voxels from sparse observations and scalability of methods become more challenging. One simply assumption is that a tensor, which has not a globally low-rank structure, can be separated into smaller block tensors without overlapping and some block tensors may has a low-rank structure (locally low-rank). Hence, we can apply tensor completion to block tensors independently, and this strategy is also useful to handling a large-scale tensor. For HaLRTC, the tuning parameters were carefully selected to yield the best possible performances. BTC methods as well as iHOOI can automatically adapt tensor rank to data. These methods were applied to MRI data completion with varying missing ratios and noise conditions. As shown in Table~\ref{tab:MRIresults}, BTC-T/L achieve the best performance under all different conditions. When missing ratio is low (e.g., 50\%), HaLRTC outperforms iHOOI, while iHOOI outperforms HaLRTC when missing ratio is larger than 60\%. Since high quality of MRI images is required for medical diagnosis, higher missing ratios (e.g. $>$ 90\%) are not tested. In addition, we also applied these methods to MRI data globally for comparisons. \mbox{WTucker} was also used for evaluations, while its tensor rank was specified as the inferred rank by BTC method. As shown in Table~\ref{tab:MRIresults2}, BTC always outperforms the other methods under different missing ratios. Note that WTucker can possibly obtain better performance than iHOOI and HaLRTC when tensor rank was specified by inferred rank from BTC under 80\% missing ratio. From  Table~\ref{tab:MRIresults}, \ref{tab:MRIresults2}, we can observe that when missing ratio is large, global tensor completion yields better performance than local tensor completion strategy. However, global tensor completions are limited by data size. The visual quality is shown in Fig.~\ref{fig:MRIvisual} which was obtained by BTC-T method.

\begin{table*}
\renewcommand{\arraystretch}{1.1}
\caption{\small The performance of MRI completion evaluated by PSNR and RRSE. For noisy MRI, the standard derivation of Gaussian noise is 3\% of brightest tissue. MRI tensor  is of size $181\times 217\times 165$ and each block tensor is of size $50\times 50\times 10$.}
\label{tab:MRIresults}
\centering
%\resizebox{1\textwidth}{!}
%{
\begin{tabular}{ c |c | c | c | c | c | c | c | c}
\hline
  & \multicolumn{2}{c|}{50\%}  & \multicolumn{2}{c|}{60\%} & \multicolumn{2}{c|}{70\%}  & \multicolumn{2}{c}{80\%} \\
  \cline{2-9}
  & Original & Noisy  & Original & Noisy  & Original & Noisy & Original & Noisy \\
 \hline
BTC-T &  27.32\quad 0.11 & 26.18\quad0.12  & 25.30\quad 0.14 & 24.60\quad 0.15  & 22.81\quad 0.18 & 22.35\quad 0.19 &20.14\quad 0.25 &20.00\quad 0.25 \\
BTC-L &  26.91\quad 0.11 & 25.57\quad0.13  & 24.84\quad 0.15 & 23.95\quad 0.16  & 22.76\quad 0.19 & 22.09\quad 0.20 &20.12\quad 0.25 &19.80\quad 0.26 \\
iHOOI & 22.69\quad 0.19 & 21.45\quad 0.22 & 22.47\quad 0.19 & 21.16\quad 0.22 & 21.63\quad 0.21 & 20.11\quad 0.25 & 18.65\quad 0.30& 17.89\quad 0.32 \\
HaLRTC & 24.84\quad 0.15 & 23.60\quad 0.17  & 22.35\quad 0.19 & 21.65\quad 0.21  & 19.93\quad 0.26 & 19.55\quad 0.27 &17.37\quad 0.34 &17.15\quad 0.35\\
%WTucker & 22.26\quad 0.20 & 19.42\quad 0.27 & 24.93\quad 0.14 & 14.10\quad 0.50 & 20.43\quad 0.24 & 18.55\quad 0.30 && \\
\hline
\end{tabular}
%}
\end{table*}

\begin{table}
\renewcommand{\arraystretch}{1.1}
\caption{\small Global tensor completion for MRI data. }
\label{tab:MRIresults2}
\centering
%\resizebox{1\textwidth}{!}
%{
\begin{tabular}{ c |c | c | c | c }
\hline
  & \multicolumn{2}{c|}{50\%}  & \multicolumn{2}{c}{80\%} \\
  \cline{2-5}
  & PSNR  &  RRSE & PSNR  &  RRSE  \\
 \hline
BTC-T &  26.40  & 0.12  & 22.33 &  0.19    \\
WTucker & 19.42  &  0.27 & 21.53  & 0.21   \\
iHOOI & 21.57  &  0.21 & 19.77  & 0.26   \\
HaLRTC & 23.12  &  0.18 & 17.06  & 0.36   \\
\hline
\end{tabular}
%}
\end{table}

\section{Conclusion}
\label{sec:conclusion}
We have proposed a class of Bayesian Tucker models for both tensor decomposition and completion. To modeling structural sparsity, we introduce two group sparsity inducing priors by the hierarchical representations. The model inferences, especially for the non-conjugate Laplace priors, are derived under variational Bayesian framework. Our models can infer an optimal multilinear rank from whether a fully or partially observed tensor by automatical model reduction, yielding significant advantages for tensor completion. For algorithm implementation, we propose several Theorems related to multilinear operations to improve computational efficiency and scalability. Empirical results on synthetic data as well as chemometrics and neuroscience applications validated the superiority of our models in terms of tensor decomposition and completion.

%\section*{Acknowledgments}

\bibliographystyle{IEEEtran}
\bibliography{IEEEabrv,BayesTucker}

\begin{IEEEbiography}[{\includegraphics[width=1in,height=1.25in,clip,keepaspectratio]{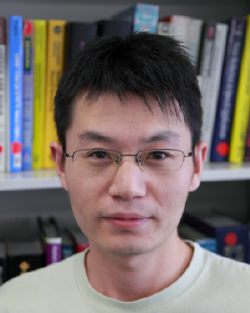}}]{Qibin Zhao} received the Ph.D. degree from Department of Computer Science and Engineering, Shanghai Jiao Tong University, Shanghai, China, in 2009. He is currently a research scientist at Laboratory for Advanced Brain Signal Processing in RIKEN Brain Science Institute, Japan and is also a visiting professor in Saitama Institute of Technology, Japan. His research interests include machine learning, tensor factorization, computer vision and brain computer interface. He has published more than 50 papers in international journals and conferences.
\end{IEEEbiography}

\vspace{-0in}
\begin{IEEEbiography}[{\includegraphics[width=1in,height=1.25in,clip,keepaspectratio]{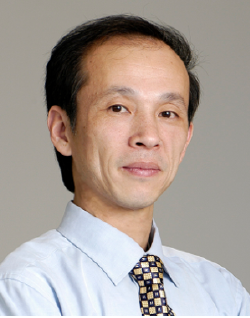}}]{Liqing Zhang} received the Ph.D. degree from Zhongshan University, Guangzhou, China, in 1988. He is now a Professor with Department of Computer Science and Engineering, Shanghai Jiao Tong University, Shanghai, China. His current research interests cover computational theory for cortical networks, visual perception and computational cognition, statistical learning and inference. He has published more than 210 papers in international journals and conferences.
\end{IEEEbiography}

\vspace{-0in}
\begin{IEEEbiography}[{\includegraphics[width=1in,height=1.25in,clip,keepaspectratio]{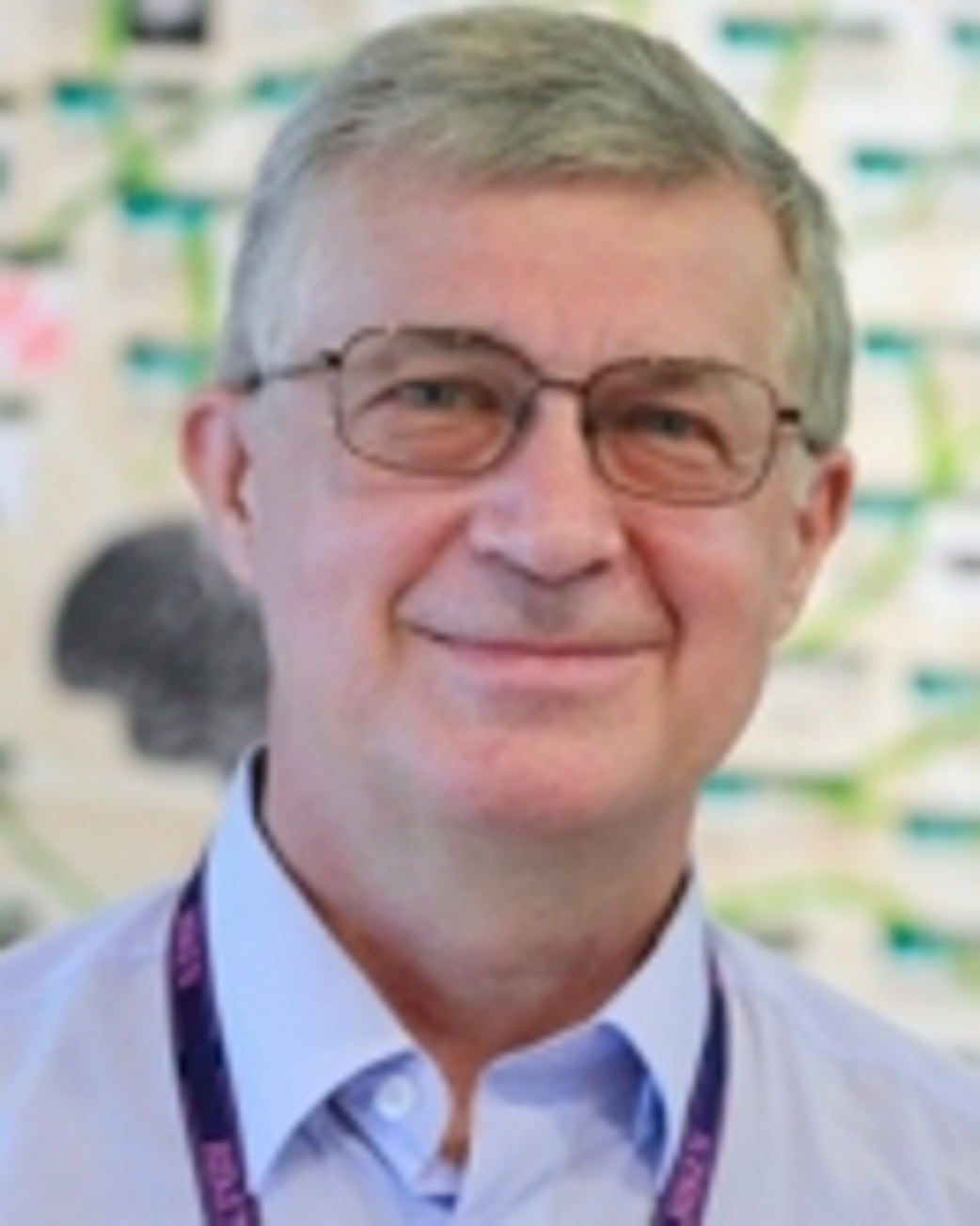}}]{Andrzej Cichocki} received the Ph.D. and Dr.Sc. (Habilitation) degrees, all in electrical engineering, from Warsaw University of Technology (Poland). He is the senior team leader of the Laboratory for Advanced Brain Signal Processing, at RIKEN BSI (Japan). He is coauthor of more than 400 scientific papers and 4 monographs (two of them translated to Chinese). He served as AE of IEEE Trans. on Signal Processing, TNNLS, Cybernetics and J. of Neuroscience Methods.
\end{IEEEbiography}
\vfill

%\enlargethispage{-5in}

\end{document}

%% file: DefinedComand.tex
\newcommand{\tensor}[1]{\boldsymbol{\mathcal{#1}}}

\theoremstyle{plain}
\newtheorem{theorem}{Theorem}[section]

\theoremstyle{definition}
\newtheorem{definition}[theorem]{Definition}
\theoremstyle{remark}